\documentclass[letterpaper]{article} 
\usepackage{aaai2026}  
\usepackage{times}  
\usepackage{helvet}  
\usepackage{courier}  
\usepackage[hyphens]{url}  
\usepackage{graphicx} 
\usepackage{natbib}  
\usepackage{caption} 
\usepackage{algorithm}
\usepackage{algorithmic}

\usepackage{newfloat}
\usepackage{listings}
\DeclareCaptionStyle{ruled}{labelfont=normalfont,labelsep=colon,strut=off} 
\floatstyle{ruled}
\newfloat{listing}{tb}{lst}{}
\floatname{listing}{Listing}
\usepackage{threeparttable}
\usepackage{multirow}
\usepackage{booktabs}
\usepackage{colortbl} 
\usepackage{xcolor}
\usepackage{tabularx}
\usepackage{amsmath}
\usepackage{amssymb}
\usepackage{dsfont}
\usepackage{epstopdf}

\title{Boosting Adversarial Transferability via Ensemble Non-Attention}
\author {
	Yipeng Zou\textsuperscript{\rm 1},
	Qin Liu\textsuperscript{\rm 1}\thanks{Corresponding author.},
	Jie Wu\textsuperscript{\rm 2, \rm 3}, 
	Yu Peng\textsuperscript{\rm 4},
	Guo Chen\textsuperscript{\rm 1},
	Hui Zhou\textsuperscript{\rm 1},
	Guanghui Ye\textsuperscript{\rm 1}
}
\affiliations {
	\textsuperscript{\rm 1}College of Computer Science and Electronic Engineering, Hunan University\\
	\textsuperscript{\rm 2}China Telecom Cloud Computing Research Institute\\
	\textsuperscript{\rm 3}Department of Computer and Information Sciences, Temple University\\
	\textsuperscript{\rm 4}Laboratory of Intelligent Collaborative Computing, University of Electronic 
	Science and Technology of China \\
	\{cszyp, gracelq628, guochen, huizhoucsee, yghui\}@hnu.edu.cn, jiewu@temple.edu, ypeng@uestc.edu.cn
}

\begin{document}

\maketitle

\begin{abstract}
	Ensemble attacks  integrate the outputs of surrogate models with diverse architectures,
	which can be combined with various gradient-based   attacks
	to improve adversarial  transferability.
	However, previous work shows  unsatisfactory attack  performance when transferring across heterogeneous  model architectures. 
	The main reason is that  
	the gradient update directions of heterogeneous surrogate  models
	differ widely, making  it hard to reduce the gradient
	variance of ensemble  models   while making the best of    individual    model.
	To tackle this challenge, we
	design a novel     ensemble attack,  \textsf{NAMEA},
	into the iterative gradient optimization process.
	Our design is    inspired by the  observation that
	the attention areas of   heterogeneous   models   vary sharply,
	thus the non-attention areas  of  ViTs are  likely to be the focus of CNNs and vice versa.
	Therefore, we merge  the gradients respectively  from the attention and non-attention areas of   ensemble models so as
	to  fuse the transfer information of CNNs  and ViTs.
	Specifically, we pioneer a new way of decoupling
	the gradients of non-attention areas from those of attention areas, while
	merging gradients by   meta-learning.
	Empirical evaluations on ImageNet dataset indicate   that \textsf{NAMEA} outperforms AdaEA and SMER, the state-of-the-art   ensemble attacks by
	an average of   $15.0\%$ and $ 9.6\% $, respectively.
	This  work is the first attempt to explore the power of  \emph{ensemble non-attention} in boosting cross-architecture transferability,
	providing new insights into launching ensemble attacks.
\end{abstract}


\section{Introduction}

Deep neural networks (DNNs) including
convolutional neural networks (CNNs) and vision transformers (ViTs)~\cite{resenet,vitb}
are found to be  highly vulnerable to adversarial examples.
Worse still, adversarial examples crafted from surrogate models  are transferable to
unknown target models, making black-box attacks
feasible in real-world applications.
To better understand the vulnerabilities of   DNNs,
various  transferability enhancement approaches have been proposed~\cite{sim2019,DIM2019}.
Thereinto, ensemble
attacks that integrate the predictions, losses,
or logits of   surrogate models to calculate the gradients with regard  to updating adversarial examples,
have shown superior   adversarial transferability as they can mislead  multiple surrogate models at once~\cite{ED}.

\begin{figure}[!t]
	\centering
	\includegraphics[scale=0.31]{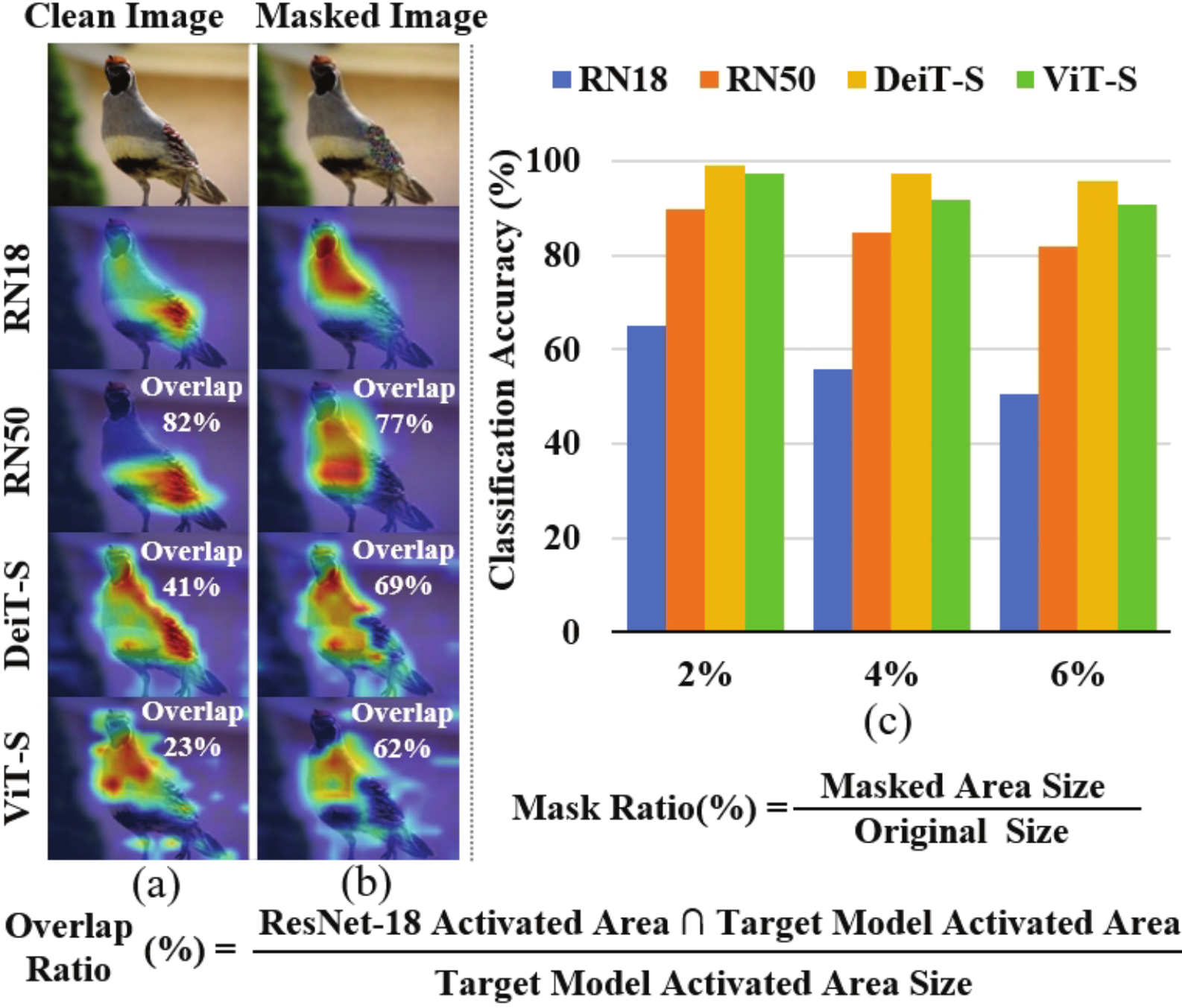}
	\caption{Attention heatmaps and classification accuracies of  clean   and    masked images. A masked image is crafted by replacing the attention area of ResNet-18 with random noises.
		Target models include ResNet-50, DeiT-S, and ViT-S.
	} \label{fig1}
\end{figure}

However, previous work
mainly focused on  transferring
across models  with homogeneous architectures (e.g., from  surrogate CNNs to target CNNs ),
exhibiting poor performance when transferring across heterogeneous model architectures (e.g., from surrogate CNNs and ViTs     to    target CNNs and ViTs).
The root cause   is that  
the gradient update directions of heterogeneous surrogate  models differ widely.
For this reason, even the  state-of-the-art (SOTA) ensemble attacks
found  it hard to balance between reducing the  gradient
variance of ensemble  models  and making the best of    individual    model, thus
easily falling  into local optimality. For instance, AdaEA~\cite{AdaEa}   mitigated gradient variance across surrogate models
by  a discrepancy-reducing filter, which ensured  stable update directions at the expense of model diversity;
While  SMER~\cite{SMER}
independently optimized individual surrogate model without considering smoothing gradients, which
may cause the     attack 
optimization direction   to be less accurate.
\emph{Hence, the main challenge  lies in    how to make the best of    individual    model
	while stabilizing  update direction among  ensemble  models.}

To tackle this challenge, we propose
a \underline{n}on-\underline{a}ttention enhanced
\underline{m}eta  \underline{e}nsemble \underline{a}ttack, \textsf{NAMEA}. 
Our design is inspired by the observation that  homogeneous models share many  attention areas,
but heterogeneous models focus on fairly different  areas as shown in Fig.~\ref{fig1}(a).
That is,  
the non-attention areas of CNNs are probably  to be the focus of ViTs, and vice
versa.
This observation is also  quantitatively supported by Fig.~\ref{fig1}(c), which shows
the classification accuracies of 1,000 random ImageNet   images
after masking varying  attention areas  of ResNet-18.
From this figure, we can see that
as the mask  ratio  increases,
the classification accuracies on CNNs  decline substantially (up to $30\%$),
but for   ViTs,  the accuracies drop slightly  (within $10\%$).
Meanwhile,   we
were surprised to observe that
the masked image
induced   high   ratios of attention overlaps across both  homogeneous and  heterogeneous   models as shown in Fig.~\ref{fig1}(b).
So we have a hypothesis that
cross-architecture transferability may can be   improved by  harnessing   \textbf{the non-attention areas of ensemble models}
i.e.,   \textbf{ensemble   non-attention}.

To  verify this hypothesis,
we  pioneer a new way of   decoupling     the gradients 
of  ensemble non-attention       from those of  the attention areas   of ensemble models,
while  incorporating meta-learning~\cite{mgaa} into
Our  meta-gradient optimization  method
consists of three steps: \textcircled{1}
\emph{Attention Meta-Training} that iteratively updates  gradients based on the attention areas of ensemble models.
\textcircled{2} \emph{Non-Attention Meta-Testing}  that iteratively optimizes  gradients based on the non-attention areas of ensemble models.
\textcircled{3}  \emph{Final Update} that merges gradients calculated from both the meta-training and meta-testing steps.
The first two steps  encourage obtaining diverse gradients    from  ensemble models, while
the last step  aims to  find a  balance  between stable update direction and model diversity.
Especially, we construct a non-attention extraction (NAE)     module    based on Grad-CAM~\cite{CAM}   to extract   (non-)attention areas,
while    designing      a gradient scaling  optimization (GSO)
module to   boost adversarial transferability in   meta-testing  step.

\begin{figure}[!t]
	\centering
	\includegraphics[scale=0.215]{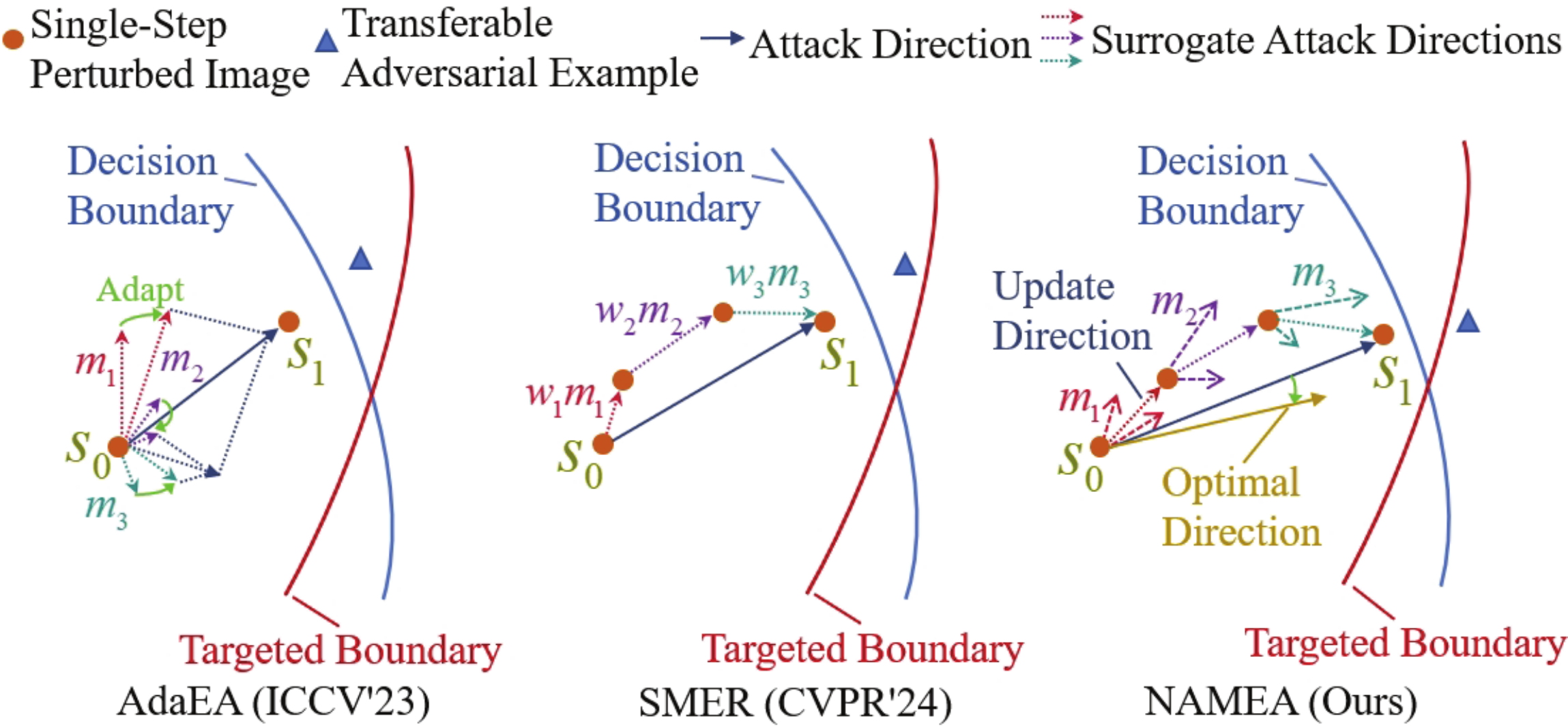}
	\caption{The attack direction search strategies of AdaEA, SMER and \textsf{NAMEA}.
		AdaEA focuses on reducing gradient discrepancies to improve attack effectiveness. SMER leverages model diversity to search the   attack direction.
		\textsf{NAMEA} merges gradients of attention and non-attention areas by meta-learning to obtain a   more accurate attack direction. }  \label{fig2}
\end{figure}

It is worth noticing that \emph{while
	attention mechanism or meta-learning had been   employed in   adversarial attacks~\cite{attention_sa,ATA},
	prior work  focused on   transferring
	across homogeneous models
	without considering the large gradient differences  in heterogeneous ensemble  models.}
In contrast,
\textbf{our work is the first   to  put forward the concept of   ensemble non-attention,
	while  merging  gradients by meta learning,
	thus tackling the
	core challenge in  improving cross-architecture transferability.}
The major  differences from the SOTA ensemble attacks are shown in Fig.~\ref{fig2},
and our   contributions are summarized  as follows:
\begin{itemize}
	\item
	We propose a novel ensemble attack, \textsf{NAMEA}, which ensures    stable
	update direction   and model diversity at once,
	exhibiting superior cross-architecture transferability.  

	
	\item
	\textsf{NAMEA} innovatively  decouples  the gradients   of  ensemble non-attention  
	from  those of attention   areas of  ensemble     models, while
	incorporating  meta-learning into
	iterative gradient    optimization process for gradient merging.

	
	\item
	As a plug-and-play method,   \textsf{NAMEA}   largely enhances  ensemble attack performance,
	when combined with various gradient-based    attacks.
	Especially for ImageNet dataset, \textsf{NAMEA} outperforms SOTA ensemble attacks, AdaEA  and SMER, by an average of  $15.0\% $  and  $9.6\%$,  respectively.
	\emph{With these encouraging results, we confirm that,
		ensemble non-attention contributes to
		boosting cross-architecture transferability, and
		our  \textsf{NAMEA}    provides new insights into launching ensemble attacks.}
\end{itemize}

\section{Related Work}

This section   introduces the most relevant work
while putting the details of
adversarial attacks and defenses  into APPX. A.

\textbf{Ensemble  Attacks.}
Ens~\cite{ens} directly averaged the ensemble models' predictions to obtain an ensemble loss before launching gradient-based attacks.
\cite{ED} further introduced the logits-based ensemble losses  to enhance the adversarial transferability.
SVRE~\cite{Svre} reduced the gradient variance by 
using stochastic variance-reduced  gradients.
To transfer across CNNs and ViTs,
AdaEA~\cite{AdaEa} adaptively fused model outputs 
by monitoring and adjusting gradient contributions.
CWA~\cite{CWA} improved adversarial transferability   by adjusting the flatness of the loss function.
SMER~\cite{SMER}   emphasized the diversity of surrogate models, and introduced
ensemble reweighing  to refine ensemble weights  based on reinforcement learning.
CSA~\cite{CSA} leveraged multiple
checkpoints from a single model's training trajectory to improve transferability.

\textbf{Attention-based or Meta-Learning-based Attacks.}
For attention-based attacks,
AGTA~\cite{attention_map} guided perturbations by computing attention weights to disrupt critical features shared across CNNs.
AoA~\cite{aoa} improved transferability by 
aligning perturbations with important attention areas.
Attention-SA~\cite{attention_sa}  designed a semantic-aware attention module to guide perturbations in attention areas.
But existing    methods  target homogeneous models 
and focus on perturbing the  attention areas,
ignoring the potential    of non-attention areas in improving cross-architecture transferability.
As for meta-learning-based attacks,
MGAA~\cite{mgaa} leveraged meta-learning to simulate white-box and black-box attacks. 
LLTA~\cite{LLA} used meta-learning to train perturbations over augmented tasks, simulating cross-task attack adaptation.
MTA~\cite{mta2023} trained a meta-surrogate model to simulate  adaptation across attack tasks.
However, existing   methods normally  leverage meta-learning to reduce gradient discrepancies,
ignoring  the gradient diversity among  heterogeneous ensemble models.
\emph{In summary, our \textsf{NAMEA}
	is the first attempt to decouple the gradients of ensemble non-attention from those of   attention areas,
	while fusing gradients via meta-gradient optimization,
	which    boosts   cross-architecture transferability from a new  perspective.}

\section{Methodology}
\subsection{Preliminaries}
\textbf{I-FGSM-based Ensemble Attacks.}
Given a target model $f: X\rightarrow Y$ and a clean   image $x \in X$ with ground-truth label $y \in Y$,
an adversarial example is crafted as $x_{adv}=x+\delta$, which
fools the target model $f(x_{adv})\neq y$, where
$\delta$  is a small perturbation
constrained by $l_{\infty}$ norm~\cite{ED}.   
The optimization problem can be formally formulated as:
\begin{equation}\label{eq1}
	\arg \max_{x_{adv}} \mathcal{L}(x_{adv}, y), s.t. \|x_{adv}-x\|_{\infty}\leq \epsilon,
\end{equation}
where $\epsilon$ is the perturbation budget, and  $\mathcal{L}$ is often the cross-entropy loss.
Let $T$ and $\alpha$ be the number of iterations and the step size,
respectively. To  solve the optimization problem in Eq.~(\ref{eq1}),
I-FGSM~\cite{IFGSM} initializes  the adversarial example with clean image, i.e., $x^0_{adv} = x$
and performs iterative updates  as follows:
\begin{equation}\label{I-FGSM}
	x^{t+1}_{adv}=\mathrm{Clip}_{\epsilon}^{x}(x^t_{adv} + \alpha \operatorname{sign} (g^{t+1})),
\end{equation}
where  $\mathrm{Clip}_{\epsilon}^{x}(\cdot)$ denotes clipping the perturbation within an $\epsilon$-ball centered at the original image $x$,
$\operatorname{sign}(\cdot)$ is the sign function, and $g^{t+1}=\nabla_{x^t_{adv}} \mathcal{L}(x^t_{adv}, y)$
denotes the gradient of the loss function with respect to $x^t_{adv}$.
As  the gradients of target models are inaccessible,
ensemble attacks  craft  adversarial examples
from multiple surrogate models $\{f_1,\ldots, f_{N}\}$,
where
the gradients can  be calculated from ensemble predications, logits, or losses~\cite{ens}.



\textbf{Attention Extraction.}
Given a surrogate   model $f_n$
and an image $x$ with label $y$,
we apply Grad-CAM~\cite{CAM} to derive   $f_n$'s attention map on $x$,
denoted by $\mathbf{H}_n(x)$.
Let $A^c_{l}$ denote the $c$-th feature map in the  $l$-th layer of model $f_n$, and let
$A^c_{l}[i,j]$  be the output of the neuron with the spatial position $[i, j]$.
The importance weight  of feature map $A^c_{l}$   can be approximated with spatially pooled gradients:
\begin{equation} \label{eq:attention_map}
	\alpha_{l}^{c} =\frac{1}{Z} \sum_{i} \sum_{j} \frac{\partial f_n(x)[y]}{\partial A_{l}^{c}[i, j]},
\end{equation}
where $Z$ is a normalizing constant such that $\alpha_{l}^{c} \in [-1, 1]$ and
$f_n(x)[y]$  is the logits  of   label $y$  when feeding model $f_n$ with input $x$.
Then, $\mathbf{H}^l_n(x)$, the attention map at the $l$-th layer of model $f_n$
can be derived  by performing  $\text{ReLU}$ on  the weighted combination of  feature maps:
\begin{equation}~\label{attentionmap}
	\mathbf{H}^l_n(x)=\operatorname{ReLU}\left(\sum_{c} \alpha_{l}^{c} \cdot A_{l}^{c}\right),
\end{equation}
where $\text{ReLU}(\cdot)$ is applied  to discard negative   pixels in the attention map,
while retaining the features that support   label $y$. Therefore, the  attention map highlights the spatial regions  most relevant to   model  decision.
As the size of the
feature maps varies across different layers and models,
$\mathbf{H}^l_n(x)$ will be
upsampled back to the size of the original image using bilinear interpolation.
In this paper, a single  layer $l$ is chosen for attention extraction,
and thus $\mathbf{H}_n(x)=\mathbf{H}^l_n(x)$.


\begin{figure*}[!t]
	\centering
	\includegraphics[scale=0.18]{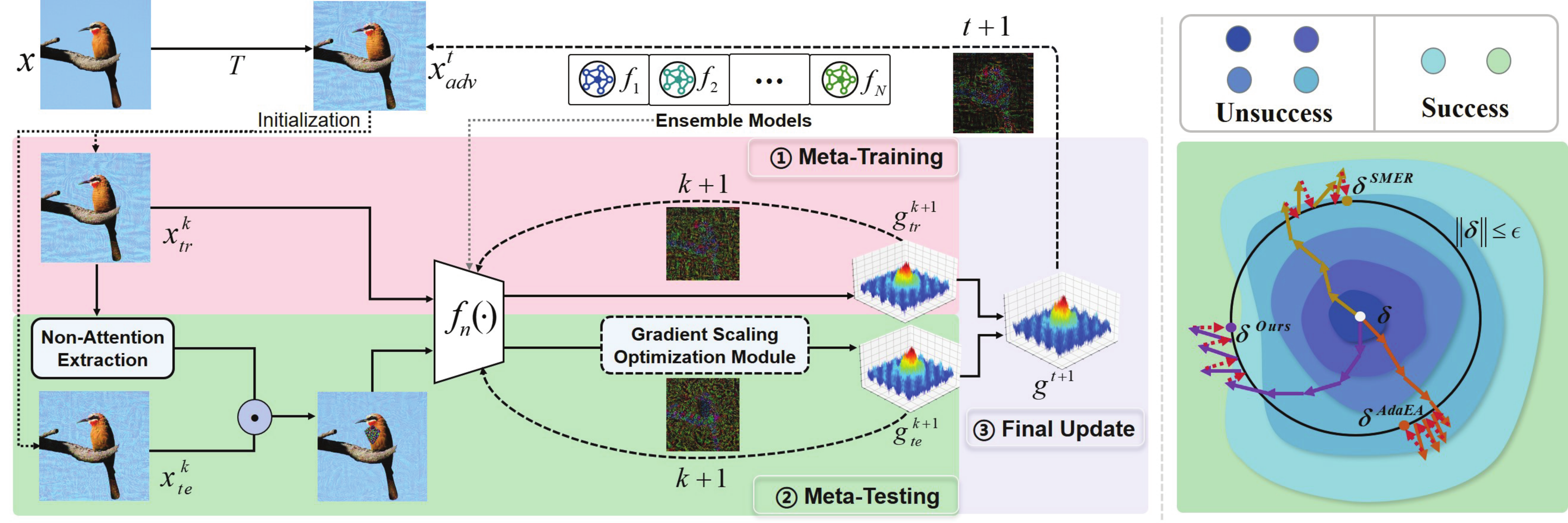}
	\caption{ Overview of  \textsf{NAMEA}. \textbf{Left:} Meta-gradient optimization process.
		Attention meta-training   updates the gradient $g^{k+1}_{tr}$ based on model's attention areas;
		Non-attention meta-testing    updates the  gradient $g^{k+1}_{te}$   based on model's  non-attention areas;
		Final update   merges the gradients from   meta-training and meta-testing steps to obtain the final gradient $g^{t+1}$.
		\textbf{Right:} The  comparison of perturbation search process.  \textsf{NAMEA} can  quickly find the   optimal   direction, avoiding falling  into local optimality.
	} \label{main}
\end{figure*}


\subsection{Meta-Gradient Optimization}
Following previous work~\cite{Svre,SMER},  \textsf{NAMEA} treats   the  iterative ensemble attack as a
stochastic gradient descent optimization process, which consists of
$T$ outer iterations
and $K$ inner loops as shown in the left side of Fig.~\ref{main}.
At a high-level view, 
each outer iteration $t$ invokes  $K$ inner loops
to calculate the   optimal  meta-training  gradient $g^K_{tr}$ and
the optimal  meta-testing  gradient $g^K_{te}$, while using  meta-learning
to obtain the merged gradient    $g^{t+1}$.
Specifically,
each inner loop $k$ picks a  random    model $f_n$ from $N$    surrogate models to extract the (non-)attention areas 
and performs a one-step update,
ensuring each surrogate model is selected at least once in every $N$ consecutive inner iterations.
From the right side of Fig.~\ref{main}, we can see that
our merged gradients fuse   the  characteristics of   ViTs and CNNs,
allowing for better model diversity  than AdaEA  
and enabling more stable   update direction than SMER.

Given $N$    surrogates  $\Theta=\{f_1,\ldots, f_N\}$ and
the adversarial example $x^t_{adv}$ at the $t$-th outer iteration,
meta-gradient optimization calculates  the merged gradient  $g^{t+1}$ as follows: 

\textbf{\textcircled{1} Attention Meta-Training.}
This step aims to  find the optimal  meta-training  gradient $g^K_{tr}$ based
on the attention areas of   selected surrogate models by running $K$ inner loops.  
The  meta-training adversarial example is    initialized as $x_{tr}^{0} = x^t_{adv}$,
and the meta-training gradient $g^{k+1}_{tr}$  and  adversarial example $x_{tr}^{k+1}$   can be   iteratively  calculated as follows:
\begin{align}\label{eq:g_train}
	g^{k+1}_{tr} &= \nabla_{x^{k}_{tr}} \mathcal{L}(x^{k}_{tr}, y),\\
	x_{tr}^{k+1}&=\mathrm{Clip}_{\epsilon}^{x}\left(x^{k}_{tr}+\alpha \operatorname{sign} ({g}^{k+1}_{tr})\right),
\end{align}
where $\mathcal{L}(x^{k}_{tr},y)=-\mathbf{1}_y\cdot \log(\operatorname{softmax} (l_n(x^k_{tr})))$
with $\mathbf{1}_y$ being the one-hot encoding of   ground-truth label $y$,
and $l_n$   the logits of the   surrogate model $f_n$   selected at inner loop $k$.

\textbf{\textcircled{2} Non-Attention Meta-Testing.}
This step aims to find the optimal meta-testing gradient $g^K_{te}$ based on the non-attention areas of selected  surrogate models
by running $K$ inner loops.
This step initializes the meta-testing adversarial example as $x^0_{te} = x^t_{adv}$.
The main trick is to
design a non-attention extraction (NAE) module 
which  masks   the selected  models'     attention areas  on the meta-testing adversarial examples  before gradient calculation.
At the $k$-th inner loop,
given the adversarial examples $x^k_{tr}$ and   $x^k_{te}$ in  meta-training and meta-testing, respectively,
the NAE   module first generates an attention mask $\mathbb{M}_k$  for the  surrogate model  $f_n$ selected at the $k$-th inner iteration  as follows:
\begin{equation}\label{mask}
	\mathbb{M}_k[i,j] =\left\{
	\begin{aligned}
		1  &, \   \mathrm{if} \ \mathbf{H}_n(x^k_{tr})[i,j] >= \eta,\\
		0  &, \   \mathrm{otherwise},   \\
	\end{aligned}
	\right.
\end{equation}
where $ \mathbf{H}_n(x^k_{tr})$ is model $f_n$'s attention map on $x^k_{tr}$ calculated from Eq.~(\ref{eq:attention_map})-Eq.~(\ref{attentionmap}),
$\mathbf{H}_n(\cdot)[i,j]$ is the  attention value at the spatial position $[i,j]$,
and   $\eta$ is a threshold value determining  if the pixel at position $(i,j)$ of an image  is important.

Let $\overline{\mathbb{M}}_k=\mathbf{1}-\mathbb{M}_k$ be the non-attention mask.
The NAE   module
masks model $f_n$'s  attention area  on the meta-testing adversarial example $x^k_{te}$
with   random Gaussian noises:
\begin{align} \label{eq:masked_adv}
	x^k_{te}=\overline{\mathbb{M}}_{k} \odot x^k_{te} + \mathbb{M}_{k} \odot \xi, \ \ \  \xi \sim \mathcal{N}(0, 1),
\end{align}
where $\odot$ is the Hadamard product. The reason we fill random noises in the attention areas
is to  further   distract  the attention of selected models. From the experiment results shown in the right side of   Fig.~\ref{ablation_componets},
we can see that filling random noises yields higher attack success rates
compared with   simply setting  pixel values  to  0s or 1s.

After being processed by the    NAE   module,
the meta-testing adversarial example $x^k_{te}$
retains only the non-attention area of model $f_n$.
Thus, the meta-testing gradient $g^{k+1}_{te}$ and
the adversarial example $x_{te}^{k+1}$ can be calculated as:
\begin{align} \label{eq:g_test}
	g^{k+1}_{te} &= \nabla_{x^{k}_{te}} \mathcal{L}(x^{k}_{te}, y),\\
	x_{te}^{k+1}&=\mathrm{Clip}_{\epsilon}^{x}\left(x_{te}^{k}+\alpha \operatorname{sign} ({g}^{k+1}_{te})\right).
\end{align}

\textbf{\textcircled{3} Final Update.}
After obtaining the optimal gradients $g^K_{tr}$ and $g^K_{tr}$ from the meta-training and meta-testing steps separately,
the final update step   obtain the fused gradient as:
\begin{align}\label{eq:g_final}
	g^{t+1} = g^K_{tr} +  g^K_{te} \odot  \overline{\mathbb{M}}_{K}.
\end{align}
Then, the outer loop can update the adversarial sample  with Eq.~(\ref{I-FGSM}).
Note that the meta-testing gradient  $g^K_{te}$ is  masked before merging.
This is to  ensure     the transferable gradient information of  attention regions
will not be interfered with.

\subsection{Gradient Scaling Optimization Module}
Recent studies~\cite{ILA,GNS} have proven  that   the intermediate-layer features of CNNs       are more transferable,
and  the relatively small gradients in backpropagation of  ViTs  have negative   influence on   transferability.
Thus,
we design the gradient scaling optimization (GSO) module to further  optimize the meta-testing gradients.

\textbf{Layer-wise Gradient Scaling  for CNN.}
For CNNs, the GSO module
uses a scaling function to enhance the gradient contribution of intermediate layers.
Let $L$ denote the total number of layers.
The scaling factor of   layer $l$ ($l \in [L/3, 2L/3]$)  is defined as:
\begin{align}\label{eq:gcnn_factor}
	\lambda(l)=\lambda_{1}+\lambda_{2} \cdot \left(\frac{L}{l}\right),
\end{align}
where $\lambda_{1}$ controls the baseline scaling intensity, and $\lambda_{2}$ determines the magnitude of enhancement for each layer.
Therefore, the   shallower the layer, the larger the value of scaling factor.
In this way, we can magnify the meta-testing gradient at layer $l$ with the scaling factor $\lambda(l)$:  
\begin{align}\label{eq:gcnn_scaling}
	g_{te}[l] = g_{te}[l] \cdot \lambda(l).
\end{align}

\textbf{Channel-wise Gradient Scaling  for ViT.}
For ViTs,
the backpropagated gradient can be decomposed into $C$ channels,
$g_{te}=\{g_{te}[1], \cdots, g_{te}[C]\}$.
Thus, the GSO module uses a   scaling function
to reduce the   contribution of  channels   with low gradient magnitudes.
Let $\phi$ and $\sigma$ represent the mean and standard deviation of
the absolute gradient magnitudes across the $C$ channels, respectively.
If the meta-testing gradient at channel  $c$  is smaller than the average value of $C$ channels, 
we can shrink the  gradient magnitude    as:
\begin{align}\label{eq:gvit_scaling}
	{g}_{te}[c]  = {g}_{te}[c] \cdot \operatorname{tanh}(|\frac{g_{te}[c] - \phi}{\sigma } |).
\end{align}

In the Appendix, Alg. I  shows the overall procedure  of \textsf{NAMEA},
and   Fig. I  shows   the
adversarial examples crafted by \textsf{NAMEA} can further distract
models' attention  compared to  all competitors,
visualizing   the efficacy of \textsf{NAMEA}.

\begin{table*}[!t]
	\scalebox{0.77}{
		\begin{tabular}{ccccccccccccccccccccc}
			\toprule
			\multirow{2}{*}{Base}&\multirow{2}{*}{Attack}&\multicolumn{10}{c}{ViTs}&\multicolumn{9}{c}{CNNs}  \\
			\cmidrule(lr){3-12} \cmidrule(lr){13-21}
			\multirow{2}{*}{}&&ViT-B&PiT-B&CaiT-S&ViS&DeiT-B&TNT-S&LeViT&ConV&Swin-B&\cellcolor[gray]{0.9}\textbf{Avg.}&RN50&RN152&DN201&DN169&VGG16&VGG19&WRN101&BiT50&\cellcolor[gray]{0.9}\textbf{Avg.} \\ \hline
			\multirow{3}{*}{}&Ens&16.0&10.7&25.0&17.2&26.8&28.4&17.9&30.8&9.9&\cellcolor[gray]{0.9}20.3&22.7&13.0&30.3&34.7&35.5&33.6&22.6&28.2&\cellcolor[gray]{0.9}27.6 \\
			\multirow{3}{*}{}&SVRE&13.1&11.5&21.9&19.2&23.2&28.2&19.3&23.9&10.1&\cellcolor[gray]{0.9}18.9&29.0&16.2&34.8&39.5&42.1&28.9&26.0&32.5&\cellcolor[gray]{0.9}32.4  \\
			\multirow{3}{*}{I-FGSM}&AdaEA&25.1&17.6&39.2&27.5&40.4&40.2&28.8&42.7&15.6&\cellcolor[gray]{0.9}30.8&38.7&21.1&47.0&50.1&53.0&48.4&34.5&39.6&\cellcolor[gray]{0.9}41.6  \\
			\multirow{3}{*}{}&CWA&27.8&10.6&41.5&16.7&49.9&46.7&21.1&48.8&11.7&\cellcolor[gray]{0.9}30.5&12.9&6.9&20.8&22.6&34.3&32.1&15.2&25.5&\cellcolor[gray]{0.9}21.3  \\
			\multirow{3}{*}{}&SMER&27.4&16.4&42.6&26.0&43.9&44.7&27.7&48.9&15.4&\cellcolor[gray]{0.9}32.6&33.2&18.4&43.1&45.7&50.0&48.4&31.4&39.6&\cellcolor[gray]{0.9}38.7  \\
			\multirow{3}{*}{}&CSA&27.5&17.8&42.1&27.3&43.0&48.6&30.4&43.7&16.0&\cellcolor[gray]{0.9}32.9&36.6&20.4&49.7&50.2&51.9&51.0&36.2&42.3&\cellcolor[gray]{0.9}42.3  \\
			\multirow{3}{*}{}&\textbf{Ours}&\textbf{43.0}&\textbf{25.5}&\textbf{61.2}&\textbf{38.0}&\textbf{63.0}&\textbf{61.2}&\textbf{42.9}&\textbf{63.6}&\textbf{21.8}&\cellcolor[gray]{0.9}\textbf{46.7}&\textbf{46.2}&\textbf{26.4}&\textbf{55.8}&\textbf{58.5}&\textbf{64.4}&\textbf{60.7}&\textbf{43.8}&\textbf{52.1}&\cellcolor[gray]{0.9}\textbf{51.0}  \\ \hline
			\multirow{3}{*}{}&Ens&34.0&24.9&48.5&34.7&51.7&49.8&38.7&51.2&20.6&\cellcolor[gray]{0.9}39.3&43.4&26.5&52.8&53.6&55.2&52.9&39.6&46.4&\cellcolor[gray]{0.9}46.3  \\
			\multirow{3}{*}{}&SVRE&31.3&24.2&43.2&35.1&44.6&50.5&38.9&46.5&19.3&\cellcolor[gray]{0.9}37.1&49.6&30.5&58.1&60.5&59.2&58.0&45.3&50.6&\cellcolor[gray]{0.9}51.5  \\
			\multirow{3}{*}{MI-FGSM}&AdaEA&41.2&25.5&56.3&38.8&59.4&55.8&41.4&58.7&21.7&\cellcolor[gray]{0.9}44.3&49.0&29.2&56.2&59.9&59.5&57.8&43.7&52.2&\cellcolor[gray]{0.9}47.6  \\
			\multirow{3}{*}{}&CWA&35.1&18.4&53.5&28.6&55.4&56.7&38.9&58.2&18.0&\cellcolor[gray]{0.9}40.3&37.7&22.1&48.7&51.4&58.8&53.6&37.5&44.9&\cellcolor[gray]{0.9}44.3  \\
			\multirow{3}{*}{}&SMER&45.4&26.8&61.2&40.2&63.0&61.8&47.5&64.9&25.1&\cellcolor[gray]{0.9}48.4&51.0&31.5&59.8&61.0&66.0&61.7&47.9&55.1&\cellcolor[gray]{0.9}54.3  \\
			\multirow{3}{*}{}&CSA&48.5&29.8&61.3&45.4&63.2&66.2&49.0&64.2&27.1&\cellcolor[gray]{0.9}50.5&52.0&32.0&60.4&62.3&66.1&63.6&49.6&53.8&\cellcolor[gray]{0.9}55.0  \\
			\multirow{3}{*}{}&\textbf{Ours}&\textbf{56.6}&\textbf{34.9}&\textbf{72.6}&\textbf{51.1}&\textbf{74.5}&\textbf{72.5}&\textbf{59.0}&\textbf{74.5}&\textbf{32.8}&\cellcolor[gray]{0.9}\textbf{58.7}&\textbf{59.7}&\textbf{39.7}&\textbf{69.9}&\textbf{69.9}&\textbf{73.3}&\textbf{72.2}&\textbf{57.1}&\textbf{63.4}&\cellcolor[gray]{0.9}\textbf{63.2}  \\ \hline
			\multirow{3}{*}{ }&Ens&42.5&38.3&56.6&50.5&56.1&62.0&53.7&59.3&31.4&\cellcolor[gray]{0.9}50.0&59.5&41.9&70.1&71.5&71.4&70.0&60.4&63.5&\cellcolor[gray]{0.9}63.5  \\
			\multirow{3}{*}{}&SVRE&45.2&43.1&65.4&57.0&62.5&70.5&63.0&63.3&32.2&\cellcolor[gray]{0.9}55.8&66.8&49.1&76.7&77.8&78.2&75.4&67.7&71.7&\cellcolor[gray]{0.9}70.4  \\
			\multirow{3}{*}{DI-MI-FGSM}&AdaEA&47.7&36.6&67.2&52.6&66.2&69.3&56.0&66.4&30.8&\cellcolor[gray]{0.9}54.8&60.5&42.1&69.3&72.4&72.8&70.9&58.5&64.9&\cellcolor[gray]{0.9}63.9  \\
			\multirow{3}{*}{}&CWA&53.6&44.4&73.6&57.9&71.1&79.4&66.1&73.7&33.1&\cellcolor[gray]{0.9}61.4&64.3&47.8&76.7&77.9&79.6&78.5&65.0&72.9&\cellcolor[gray]{0.9}70.3 \\
			\multirow{3}{*}{}&SMER&66.9&57.2&81.9&70.6&82.0&85.4&75.7&83.2&46.0&\cellcolor[gray]{0.9}72.1&75.3&59.2&85.0&85.7&84.0&82.7&75.5&80.2&\cellcolor[gray]{0.9}78.5 \\
			\multirow{3}{*}{}&CSA&54.8&46.2&68.1&60.1&69.2&73.4&63.2&68.8&38.8&\cellcolor[gray]{0.9}60.3&63.2&48.2&76.7&75.2&76.5&73.4&66.2&72.7&\cellcolor[gray]{0.9}69.0  \\
			\multirow{3}{*}{}&\textbf{Ours}&\textbf{72.7}&\textbf{63.6}&\textbf{85.9}&\textbf{77.8}&\textbf{86.5}&\textbf{89.2}&\textbf{80.8}&\textbf{86.6}&\textbf{54.1}&\cellcolor[gray]{0.9}\textbf{77.5}&\textbf{80.9}&\textbf{68.4}&\textbf{88.6}&\textbf{89.4}&\textbf{87.7}&\textbf{87.6}&\textbf{81.1}&\textbf{86.0}&\cellcolor[gray]{0.9}\textbf{83.7}
			\\
			\bottomrule
	\end{tabular} }
	\centering
	\caption{Comparison of   ASRs ($\%$) between \textsf{NAMEA} and  baselines. For all the tables, the best results are highlighted in bold.  }  \label{tab:attack-base}
\end{table*}

\section{Experiments} \label{experi}
The attack performance is assessed on 3 benchmarks against 9  ViTs, 8 CNNs,  6 hybrid models, 6 defense models, and 9 defense methods.
For ImageNet dataset, we adopt 6 gradient-based basic attacks.
Due to limited space,  this section only presents the representative results on ImageNet dataset.
In the Appendix, we will provide the
experiment results on CIFAR-10 and CIFAR-100 datasets,
comparison of computational and memory overheads, visualization of attack performance,
and supplementary results on ImageNet  in terms of  transferability, robustness, and ablation studies.

\subsection{Experiments Setup}
\textbf{Datasets and Models.}
ImageNet~\cite{imagenet2012}, the benchmark  dataset contains 1000 categories with about  1.2 million images.
To align with previous work~\cite{TGR,PNA}, we randomly select
one   image   
from each class to form the test set.
Following~\cite{AdaEa},
we employ
ViT-T\cite{vitb}, DeiT-T~\cite{deit}, ResNet-18 (RN18)~\cite{resenet}, and Inception-v3 (Inc-v3)~\cite{incv3}
as the surrogate models. 
The  target models  
include different architectures: \textcircled{1}
ViT models~\cite{pitB,cait,visform,tnts,levit,convb,swin}:
ViT-B, PiT-B, CaiT-S, Visformer-S (ViS), DeiT-B, TNT-S, LeViT, ConViT-B (ConV), and Swin-B.
\textcircled{2}  CNN models~\cite{densenet,vgg19,widenet,kolesnikov2020big}:  RN50, RN152, DenseNet-201 (DN201),   DN169,
VGG16, VGG19, WideResNet-101 (WRN101), and BiT-M-R50$\times$1 (BiT50).
\textcircled{3}  Defense  models~\cite{incv3,resv2-152,tramer2018ensemble}: Inc-v4, Inc-RN-v2 (IR-v2), Inc-v3-adv (Inc-v3$_{adv}$), Inc-v3-ens3 (Inc-v3$_{ens3}$), Inc-v3-ens4 (Inc-v3$_{ens4}$), and  Inc-RN-v2-ens (IR-v2$_{ens}$).

\textbf{Baselines and Metrics.}
We compare the  attack success rate (ASR)
with six ensemble attacks:  Ens~\cite{ens}, SVRE~\cite{Svre}, AdaEA~\cite{AdaEa},  CWA~\cite{CWA}, SMER~\cite{SMER},
and CSA~\cite{CSA}
under the same ensemble settings and perturbation budget $\epsilon = 8/255$.
Moveover, we report the average results of 5 trials, with a  deviation of less than 0.6\%.

\textbf{Parameters Settings.}
For the baselines and our \textsf{NAMEA},
we use I-FGSM, MI-FGSM~\cite{ED}, and DI-MI-FGSM~\cite{DIM2019}
as the basic attacks.
The  hyper-parameters in the baselines follow the optimal setting in  the respective literature.
For a fair comparison,
CSA  employs  7 checking points from each surrogate model, expanding  the ensemble scale to 28  models.
We set the number of outer iterations as $T = 10$ and
the number of internal loops as $K=16$, 
using
step size   $\alpha = 0.8/255$  
and   momentum decay  $\mu = 1.0$.
Besides, ViTs   use  the output of the  pre-activation normalization layer in  final self-attention blocks, 
RN18  uses the output of the last convolutional block in   final residual stage, 
and Inc-v3 uses the output of $Mixed\_7b$ to extract attention areas,
where  the attention  threshold is set to $\eta = 0.6$.

\begin{table}[!t]
	\scalebox{0.82}{
		\centering
		\begin{tabular}{cccccccc}
			\toprule
			\multirow{2}{*}{Attack} & \multicolumn{6}{c}{Defense Models} \\
			\cmidrule(lr){2-8}
			& Inc-v4 & IR-v2 & Inc-v3$_{adv}$ & Inc-v3$_{ens3}$ & Inc-v3$_{ens4}$ & IR-v2$_{ens}$ & \cellcolor[gray]{0.9}\textbf{Avg.} \\
			\hline
			Ens & 59.6 & 65.5 & 75.1 & 49.3 & 48.9 & 56.3 & \cellcolor[gray]{0.9}59.1 \\
			SVRE & 67.6 & 69.6 & 78.1 & 53.0 & 53.5 & 59.5 & \cellcolor[gray]{0.9}63.5 \\
			AdaEA & 59.1 & 64.3 & 73.5 & 43.2 & 43.5 & 54.1 & \cellcolor[gray]{0.9}56.6 \\
			CWA & 70.9 & 71.2 & 78.2 & 50.7 & 55.1 & 59.2 & \cellcolor[gray]{0.9}64.2 \\
			SMER & 75.2 & 74.0 & 80.0 & 60.8 & 60.4 & 64.3 & \cellcolor[gray]{0.9}69.1 \\
			CSA & 64.2 & 68.1 & 77.8 & 50.7 & 52.9 & 59.2 & \cellcolor[gray]{0.9}62.1 \\
			\textbf{Ours} & \textbf{80.9} & \textbf{79.0} & \textbf{83.7} & \textbf{66.2} & \textbf{66.3} & \textbf{68.9} & \cellcolor[gray]{0.9}\textbf{74.2} \\
			\bottomrule
		\end{tabular}
	}
	\centering
	\caption{Comparison of ASRs (\%) against 6 defense models. }
	\label{tab:attack-defenseModels}
\end{table}

\subsection{Main Results}
\textbf{Cross-Architecture Transferability.}
From
Table~\ref{tab:attack-base}, we can see that our \textsf{NAMEA} achieves superior adversarial transferability, always  performing  best
when combining with different base attacks. 
While CSA achieves decent performance, 
it incurs huge time and memory costs to train  models and save checking points,
as shown in Table III of the Appendix.
But even  CSA expands
surrogate model scale with various weights,
our    \textsf{NAMEA} still works better.
And we can observe that
among all the base attacks, \textsf{NAMEA} and all baselines work  best under  DI-MI-FGSM,
followed by MI-FGSM, and  finally  I-FGSM.
The performance gain of  DI-MI-FGSM   can be attributed to
the input diversity that allows to better capture the universal   adversarial information. 
In particular, SMER shows surging attack effects under DI-MI-FGSM, because   ensemble
reweighing  makes  full use of model diversity when working together with  input diversity.
However, even  under   DI-MI-FGSM, \textsf{NAMEA} promotes the average ASR by $5.3\% $ compared to SMER.
From the supplementary  results shown in the Appendix,
we can observe that
\textsf{NAMEA} also consistently achieves the best performance under
benchmark datasets  CIFAR-10 and CIFAR-100 (Table I-II),  additional base attacks (Table IV),
hybrid target models (Table V), different perturbation budgets (Table VI), and more surrogate models (Table VII).
The above results fully  validate our hypothesis
that ensemble non-attention  contributes to improve cross-architecture transferability.

\begin{table}[!t]
	\centering
	\scalebox{0.82}{
		\begin{tabular}{ccccccccccc}
			\toprule
			\multirow{2}{*}{Attack} & \multicolumn{10}{c}{Defense Methods} \\
			\cmidrule(lr){2-11}
			&R$\&$P&HGD & NIPS-r3& JPEG & RS   & NPR & FD  & Bit-RD & DiffPure & \cellcolor[gray]{0.9}\textbf{Avg.} \\
			\hline
			{Ens} & 57.4 & 43.7 & 58.9 & 52.4 & 19.2 & 21.6 & 50.8 & 51.6 & 26.0 & \cellcolor[gray]{0.9}42.4 \\
			{SVRE}	&63.5 &51.5& 66.1    &58.6 & 19.4 & 22.1 & 55.7& 58.3 & 26.3 &\cellcolor[gray]{0.9}46.8	\\
			{AdaEA}	&56.0 &40.1& 55.1	 &53.1 & 16.8 & 17.6 & 50.5& 54.1 & 22.2 & \cellcolor[gray]{0.9}40.6 \\
			{CWA}	&65.9 &48.0& 64.1    &62.5 & 20.3 & 18.9 & 59.5& 60.8 & 26.4 & \cellcolor[gray]{0.9}47.4 \\
			{SMER}	&75.5 &61.9& 72.4    &71.3 & 24.1 & 27.0 & 68.3& 71.0 & 39.9 &\cellcolor[gray]{0.9}56.8	\\
			{CSA}	&63.8 &51.4& 63.5    &60.2 & 21.8 & 26.2 & 58.0& 60.3 & 34.9 &\cellcolor[gray]{0.9}48.9	\\
			{\textbf{Ours}}&\textbf{80.0}&\textbf{71.4}&\textbf{78.5}&\textbf{77.6} &\textbf{29.1} &\textbf{31.6} &\textbf{74.7}&\textbf{76.9} & \textbf{50.3}& \cellcolor[gray]{0.9}\textbf{63.3} \\
			\bottomrule
		\end{tabular}
	}
	\centering
	\caption{Comparison of ASRs (\%) against defense methods.
	}
	\label{tab:attack-denfenseMethods}
	
\end{table}

\begin{table}[!t]
	\fontsize{7.8}{6}\selectfont
	\scalebox{1.}{
		\begin{tabular}{ccccccccccc}
			\toprule
			\multirow{2}{*}{APIs}    &\multicolumn{5}{c}{I-FGSM} &\multicolumn{5}{c}{DI-MI-FGSM} \\
			\cmidrule(lr){2-6} \cmidrule(lr){7-11}
			&AdaEA &CWA&SMER & CSA & \textbf{Ours}&AdaEA &CWA &SMER & CSA & \textbf{Ours} \\	
			\midrule
			Google    &23&22& 24 & 24 &\textbf{30}  &43&47& 52 & 46 &\textbf{55}\\
			Alibaba   &21&20& 23 & 19 &\textbf{26}  &39&44& 48 & 43 &\textbf{53} \\
			Baidu     &29&28& 33 & 32 &\textbf{37}  &53&58& 61 & 56 &\textbf{64} \\
			\bottomrule
		\end{tabular}
	}
	\centering
	\caption{Comparing of  ASRs ($\%$) against real-world models.  }
	\label{tab:attack-real-api}
\end{table}

\textbf{Robustness of Adversarial Examples.}
We   compare the attack performance of \textsf{NAMEA} and baselines against various  defense models and defense methods under base attack DI-MI-FGSM.
Table~\ref{tab:attack-defenseModels}  shows that even for adversarially trained  models,
our \textsf{NAMEA}  consistently achieves the best transferability   among  all competitors.
Besides,
we evaluate the ASRs of  \textsf{NAMEA} and    baselines
against 9 defense methods: R$\&$P~\cite{RP}, HGD~\cite{HGD}, NIPS-r3~\cite{NIPSr3},
JPEG~\cite{guo2018countering}, RS~\cite{cohen2019certified}, NRP~\cite{NRP}, FD~\cite{liu2019feature}, Bit-RD~\cite{xu2018feature} and DiffPure~\cite{diffpure}.
The results of  Table~\ref{tab:attack-denfenseMethods} are basically
consistent with those in Table~\ref{tab:attack-defenseModels}.
Even for the powerful diffusion-model-based defense  DiffPure, our \textsf{NAMEA} outperforms baselines by 10\%,
indicating that \textsf{NAMEA}  generates highly robust adversarial examples.
The supplementary results  in Table VIII of the Appendix demonstrate that
\textsf{NAMEA} also has  the   highest robustness  among all competitors  under base attacks I-FGSM and MI-FGSM.



\textbf{Real-World Attacks.}
We locally run  \textsf{NAMEA} and five      baselines, 
and then
take the resulting adversarial examples as the inputs of  authoritative image recognition APIs, i.e., Google Vision, Alibaba Cloud, and Baidu Cloud, for inference.
Following~\cite{LLA},
we consider an attack successful if the ground-truth label of a clean sample is not present in the top-5 list of the APIs' predictions.
To  reduce deviation, we randomly select  100 adversarial examples generated by each attack for testing.
From  Table~\ref{tab:attack-real-api}, we can see    that in real-world scenarios,
\textsf{NAMEA} always performs  best among all competitors.
When we relax the success condition to  top-1 list, our average ASRs under DI-MI-FGSM
are  $60\%$, $56\%$,  and $68\%$   for   Google, Alibaba, and Baidu APIs, respectively, which are $4\%$, $3\%$,  and $5\%$ higher than the best-performing baseline  SMER.
Hence,  \textsf{NAMEA}   has superior transferability in various black-box scenarios.

\begin{figure}[!t]
	\centering
	\begin{minipage}[c]{0.23\textwidth}
		\centering
		\includegraphics[width=\linewidth]{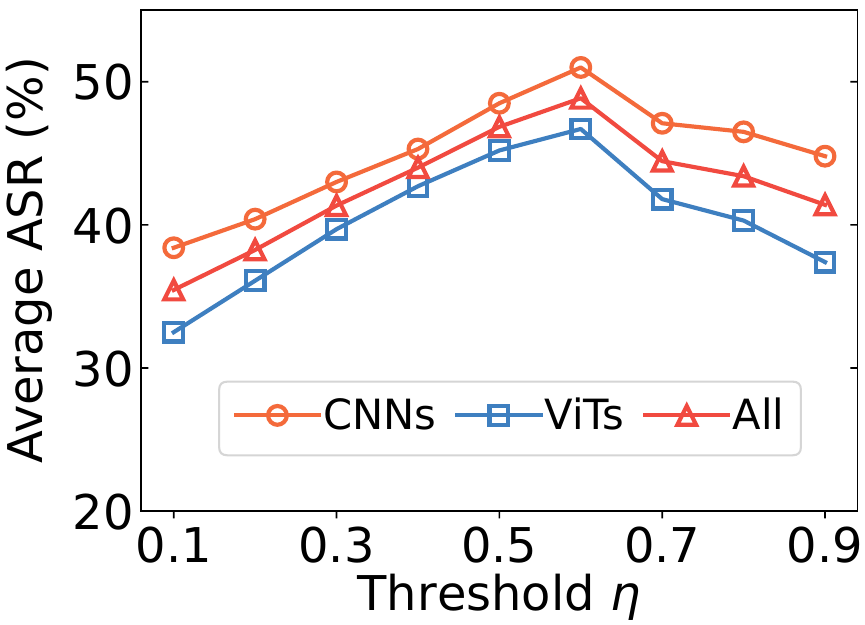}
	\end{minipage}
	\hspace{0.2em}
	\begin{minipage}[c]{0.23\textwidth}
		\centering
		\includegraphics[width=\linewidth]{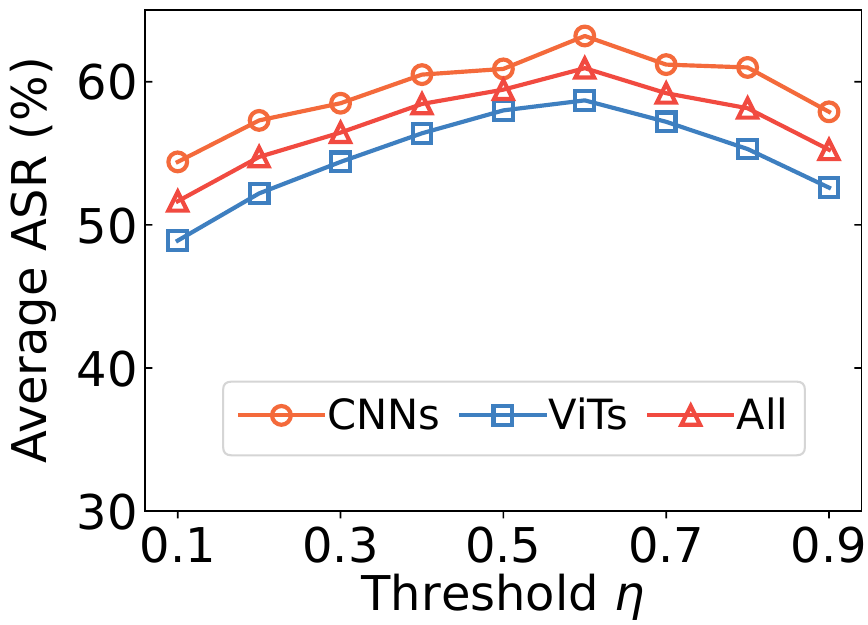}
	\end{minipage}
	\caption{Average ASRs (\%) of \textsf{NAMEA} under varying  threshold. 
		Base:     I-FGSM (\textbf{Left}) and    MI-FGSM (\textbf{Right}).}
	\label{diff_thresh}
\end{figure}

\begin{figure}[!t]
	\centering
	\begin{minipage}[c]{0.23\textwidth}
		\centering
		\includegraphics[width=\linewidth]{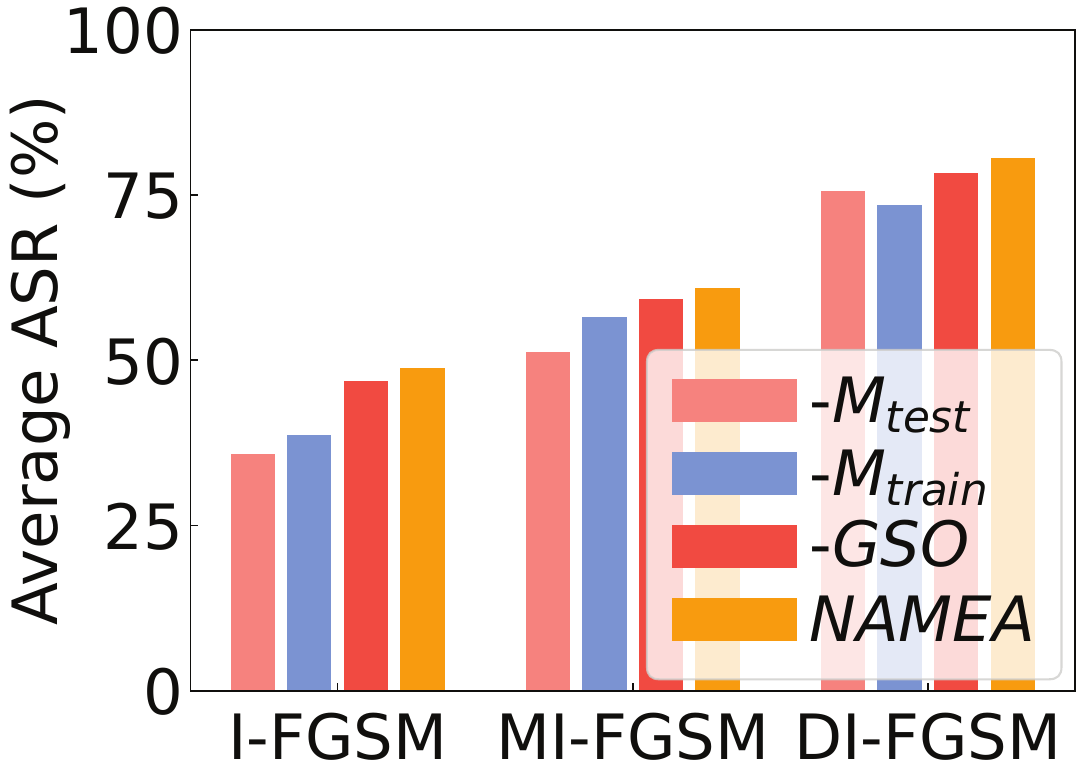}
	\end{minipage}
	\hspace{0.2em}
	\begin{minipage}[c]{0.23\textwidth}
		\centering
		\includegraphics[width=\linewidth]{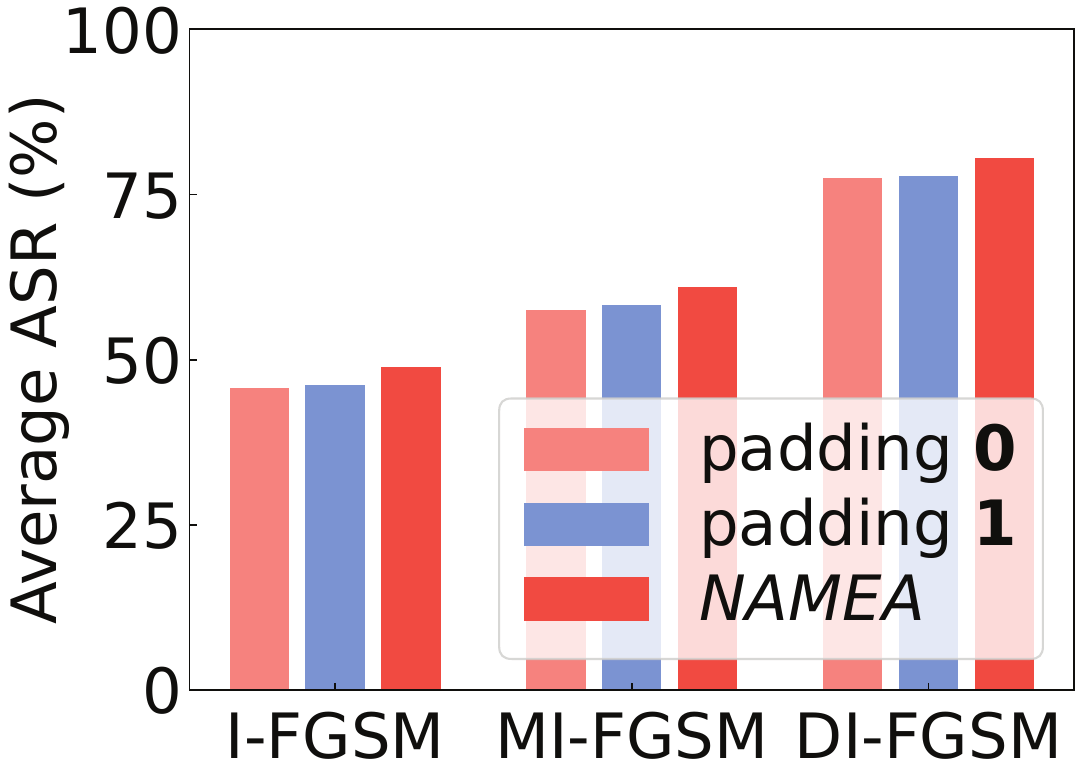}
	\end{minipage}
	\caption{\textbf{Left:} Ablation study on meta-learning and GSO.
		\textbf{Right:} Ablation study on  padding values in masked areas.}
	\label{ablation_componets}
\end{figure}

\subsection{Ablation Studies} \label{ablation_studies}
Unless otherwise specified, the ablation experiments are assessed by the average ASRs against 9 ViTs   and 8 CNNs.


\textbf{Threshold $\eta$.}
According to   Eq.~(\ref{mask}),  the smaller the value of $\eta$,
the less the number of 0s in $\mathbb{M}$,   thus the less non-attention areas being extracted.
If the value of  $\eta$ is too small,
the substantial semantic features of non-attention areas may be lost.
Hence, we need to adjust the    value to  retain basic semantics for effective exploration of non-attentive areas.
From
Fig.~\ref{diff_thresh}, we can see that when $\eta = 0.6$,  \textsf{NAMEA}  achieves the optimal result,
and the ASRs against  CNNs and ViTs show a declining trend as $\eta$ decreases or increases.
This means that  the  attack effect on CNNs and ViTs  is sensitive to  $\eta$.
We also investigate the impact of hyperparameters on CNN  gradient scaling (Eq.~(\ref{eq:gcnn_factor})) in Fig.~IV of the Appendix.

\begin{figure}[!t]
	\centering
	\begin{minipage}[c]{0.23\textwidth}
		\centering
		\includegraphics[width=\linewidth]{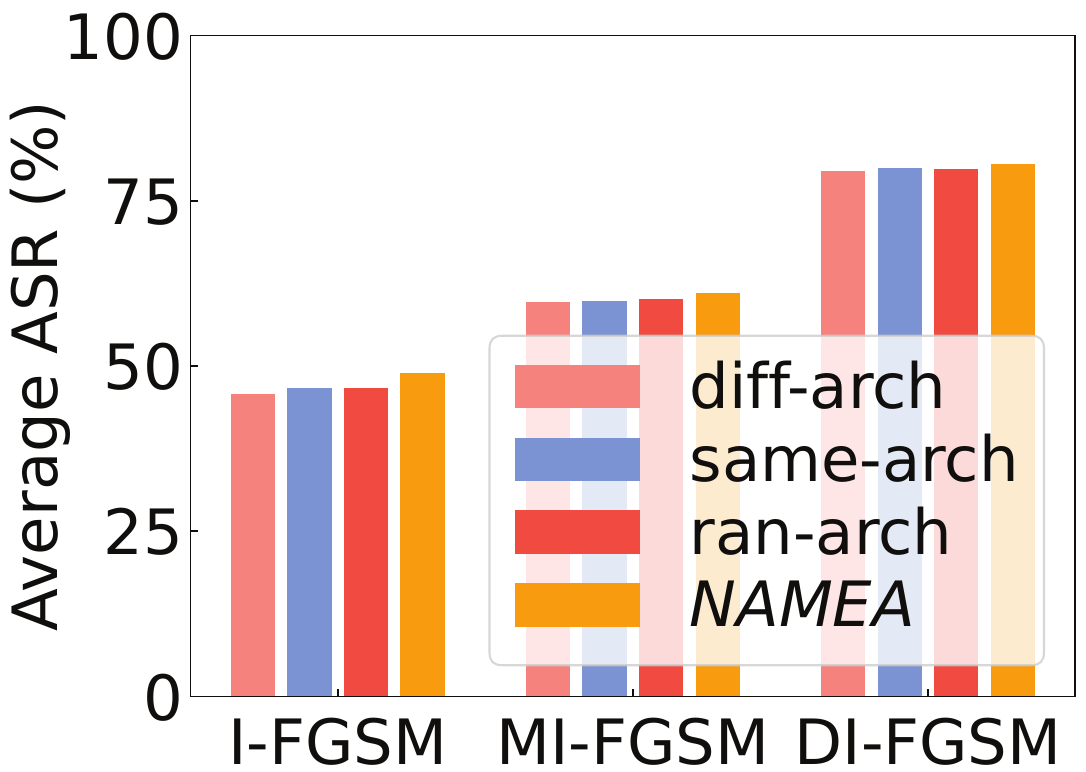}
	\end{minipage}
	\hspace{0.2em}
	\begin{minipage}[c]{0.23\textwidth}
		\centering
		\includegraphics[width=\linewidth]{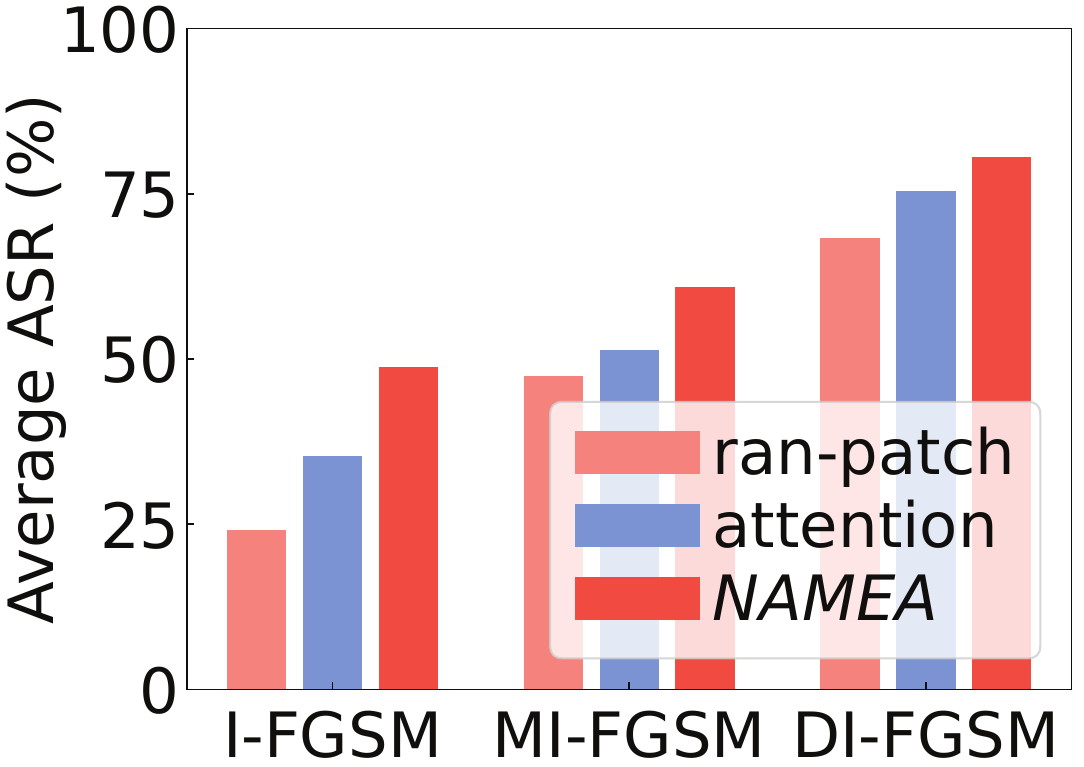}
	\end{minipage}
	\caption{\textbf{Left:} Ablation study on varying meta-testing  models.
		\textbf{Right:} Ablation study on varying meta-testing   areas.}
	\label{ablation_mask_model}
\end{figure}
\begin{figure}[!t]
	\centering
	\begin{minipage}[c]{0.23\textwidth}
		\centering
		\includegraphics[width=\linewidth]{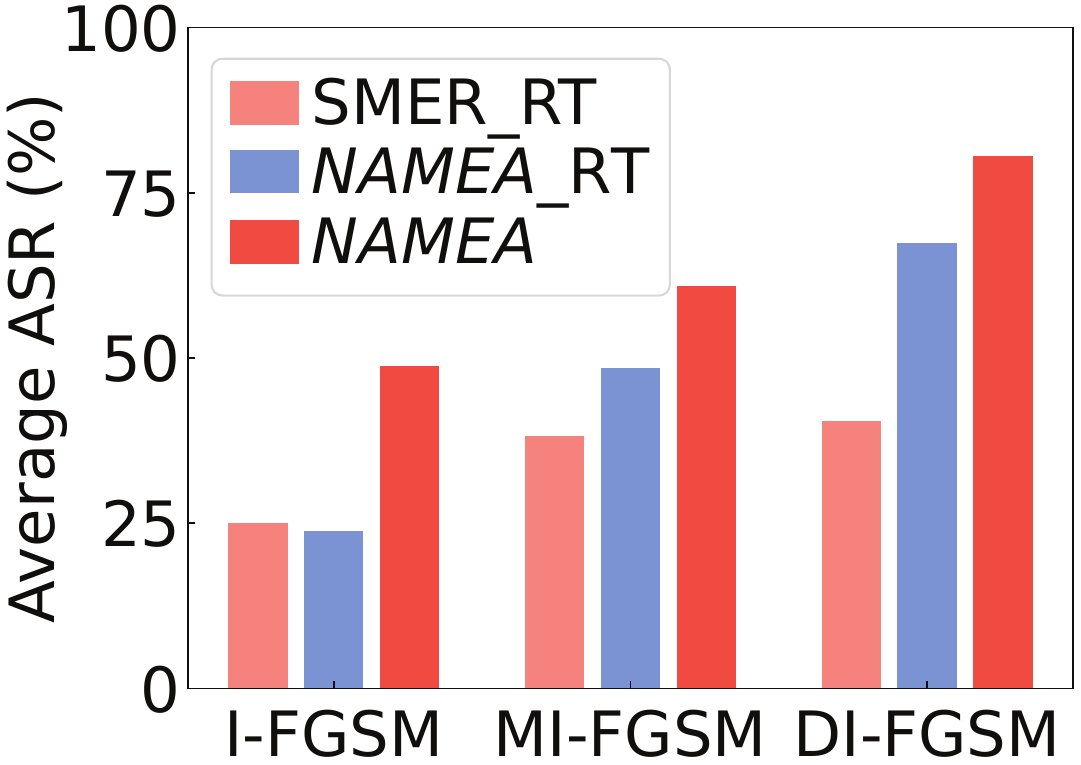}
	\end{minipage}
		\hspace{0.2em}
	\begin{minipage}[c]{0.23\textwidth}
		\centering
		\includegraphics[width=\linewidth]{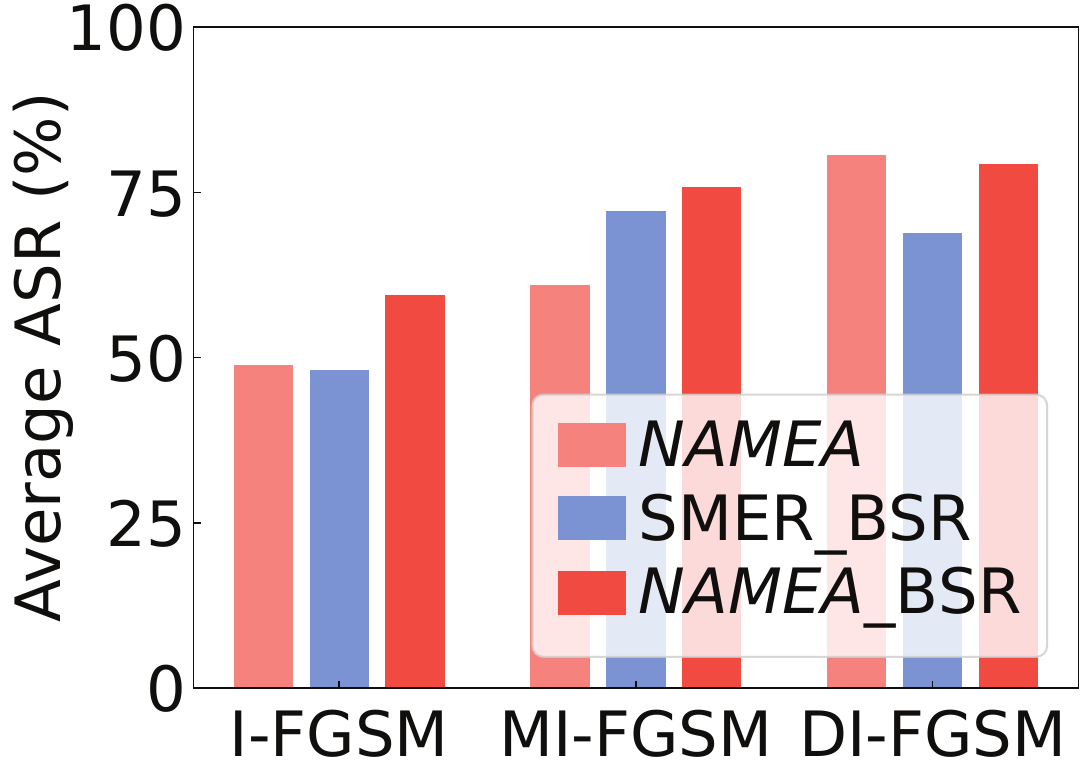}
	\end{minipage}
	\caption{Average ASRs (\%) of four comparison settings, which have  two adversarial examples in each  inner loop. }\label{ensemblena}
\end{figure}

\textbf{Meta-Learning  Steps and GSO Module.}
Let  $-M_{train}$, $-M_{test}$, and $-GSO$ denote
removing the meta-training step,   meta-testing step,   and   GSO module from  \textsf{NAMEA}, respectively.
In $-M_{train}$ and $-M_{test}$,
the final meta gradients are calculated as    $g^{t+1} =  g^K_{te} \odot  \overline{\mathbb{M}}_{K}$ and $g^{t+1} = g^K_{tr}$, respectively.
As shown in the left side of Fig.~\ref{ablation_componets}, 
if we discard the meta-testing step,
there is a   drop of 9.2\% in the average ASR;
if we  discard the meta-training step,
there is a drop of 7.4\% in the average ASR;
if we remove the GSO module, the average ASR  slightly    decreases (a drop of 2\%).
Hence
we know that the GSO module  
has  some  positive effect on    ASRs,
and   meta-learning is crucial for enhancing the  performance of \textsf{NAMEA}.
In the Appendix,   Fig. II  shows that the gradients of both attention and non-attention areas
help  craft  perturbations generalizing across CNNs and ViTs,
demonstrating the merged gradients fuse the transfer information of CNNs  and ViTs;
Fig.~III  shows that
\textsf{NAMEA} achieves the best attack performance compared with different gradient aggregation strategies and
different  gradient weights,
validating the effectiveness of meta-gradient optimization.

\textbf{Padding Values.}
According to  Eq.~(\ref{eq:masked_adv}), \textsf{NAMEA} fills
the masked areas   with random  noises.
To evaluate the impacts of different padding values on attack effect,
we evaluate the ASRs   under the other two kinds of  padding values: full $0$s and full $1$s. 
As shown in the right side of Fig.~\ref{ablation_componets},  random noises achieve the  highest   ASRs.
For instance, when using I-FGSM as the base attack, filling random noises outperform the other two filling methods by approximately 2.7$\%$ in average ASR.
This improvement may stem from the stronger disruption of   models' attention caused by random noises.

\textbf{Model  Selection Strategies.}
Meta-testing  uses the same surrogate model as meta-training.
To test the impact of different model selection strategies on   attack effect, %
we design the   ablation settings:
\textbf{diff-arch} selects a random model with   different architecture; 
\textbf{same-arch} selects a random model with the same architecture; 
\textbf{ran-arch} randomly selects a model.
As shown in the left side of Fig.~\ref{ablation_mask_model}, the average ASRs are almost unaffected by varying selection strategies.
This is because the surrogate models in the inner loop are randomly chosen,
ensuring sufficient exploration of non-attention areas of ensemble models.
In the Appendix,    
we also provide the ablation study on different ensemble settings, 
and Table IX   shows that \textsf{NAMEA} always performs best  under 
varying ratios of CNNs to ViTs and varying ensemble scales.

\textbf{Extracted Areas.}
Meta-testing  extracts  the non-attention areas from adversarial examples before gradient calculation.
To validate the critical role of non-attention areas,
we design the   ablation settings: \textbf{attention} extracts   attention areas, i.e., $x^k_{te}=\mathbb{M}_{k} \odot x^k_{te}$;
\textbf{ran-patch} extracts   random patches of    size $(56 \times 56)$.
In both settings, the final update merges the gradient as $g^{t+1} = g^K_{tr} +  g^K_{te}$.
From the right side of Fig.~\ref{ablation_mask_model}, we can see that
non-attention areas achieve the  highest   ASRs, largely  surpassing  all the other setting.
This is because non-attention areas together with attention areas can
make the best of the transferable information of individual models.



\subsection{Discussion}
\textbf{The Impact of Ensemble Non-Attention.}
\textsf{NAMEA}  
derives two  adversarial examples,
$x^k_{tr}$ and $x^k_{te}$ at each inner loop $k$,
which may create  the   illusion that the performance gain is due to diverse inputs.
To further verify the role of ensemble non-attention,
we design four comparison settings with the  same number of copies in each inner loop:
\textbf{NAMEA\_RT}  replaces the NAE module with BSR~\cite{wang2024boosting}, which
randomly transforms  $x^k_{te}$ before gradient calculation,
while merging  the   gradient  as $g^{t+1} = g^K_{tr} +  g^K_{te}$;
\textbf{SMER\_RT}  applies BSR  to generate diverse  copies for
each inner loop of  SMER and updates with  the average gradient.
\textbf{NAMEA\_BSR} and \textbf{SMER\_BSR}   directly combine    BSR  with \textsf{NAMEA} and SMER, respectively.
From  the left-side of Fig.~\ref{ensemblena}, we can see that \textsf{NAMEA} always perform best.
This is because the gradient update directions of random transformed inputs are diverse,
and  merging them directly will cause gradient conflicts.
But  
the gradients of  non-attention areas serve as a  supplement to those of attention areas,
helping to stabilize   update direction and improve model diversity.
From  the right-side of Fig.~\ref{ensemblena},
we can see that \textsf{NAMEA} 
can     fully leverage
both input and model diversities, thus yielding superior    performance.
We also observe that in DI-MI-FGSM, the ASR of  \textbf{SMER\_BSR} drops dramatically
and   that of \textbf{NAMEA\_BSR} slightly declines.
This may because the combination of two  input transformation methods
causes the inputs change too much, loosing   substantial semantic
features. 
But  \textbf{NAMEA\_BSR}  with the help of   non-attention areas
enables more stable update direction. 
Thus, we confirm that  \emph{ensemble non-attention   boosts adversarial transferability
	in a new angle different from input diversity}.

\section{Conclusion}
This work  is the first to explore the power of ensemble non-attention   in
improving cross-architecture transferability.
We propose a novel ensemble attack, \textsf{NAMEA}, which
integrates ensemble  non-attention   and  meta learning   to ensure
stable update direction and model diversity at once.
Experiment  results show that   \textsf{NAMEA} largely surpasses
the SOTA approaches, proving the validity of    ensemble non-attention.

\section*{Acknowledgments}
This work was supported by the National Natural Science Foundation
of China under Grants No. 62272150, No. 62222204, 
the Sichuan Science and Technology Program under Grants No. 2024ZDZX0011, No. 2025ZNSFSC1472,
and the Postdoctoral Fellowship Program of CPSF under Grant No. GZC20251074.

\bibliography{references}

\clearpage

\renewcommand{\thefigure}{\Roman{figure}}
\renewcommand{\thetable}{\Roman{table}}
\renewcommand{\thealgorithm}{\Roman{algorithm}}

\setcounter{secnumdepth}{1}

\setcounter{figure}{0}
\setcounter{table}{0}
\setcounter{algorithm}{0}

\appendix
\section*{Appendix}

In the Appendix, we first detail  the related work  in Section~\ref{related},
before providing the algorithm description for our \textsf{NAMEA} in Section~\ref{sec:alg}.
After
describing  evaluation results on  CIFAR-10 and  CIFAR-100  datasets in Section~\ref{sec:experiments setup},
we compare the computational and memory overheads with baselines in Section~\ref{sec:computation}.
Then, we
visualize the attack performance   in Section~\ref{sec:visualization}.
Finally, we provide supplementary results on ImageNet dataset
in terms of  transferability, robustness, and ablation studies
in Section \ref{sec:supplementary results}- Section \ref{sec:more ablations}.

\section{More Related Work}\label{related}

\subsection{Transfer-based Attacks in CNNs and ViTs}
For CNNs,
gradient optimization as a commonly used    transferability enhancement approach
uses the gradient information to iteratively maximize    loss functions.
I-FGSM~\cite{IFGSM}   iteratively updated the perturbation based on the sign of the gradient.
MI-FGSM~\cite{ED}  used a momentum term to stabilize the gradient update direction. 
Input transformation that  diversifies
input images before gradient calculation benefits   transferability.
DI-MI-FGSM~\cite{DIM2019} randomly adjusted the input image and
used the transformed image to update the   gradient of adversarial example.
TIM~\cite{TIM2019} calculated the input gradient using shifted images,
and convolved the original  gradient with a kernel matrix.
SIM~\cite{sim2019} leveraged the scale invariance of CNNs, integrating it with the gradient of the scaled copy of the original image.
BSR~\cite{wang2024boosting} split the image into blocks, which were then randomly shuffled and rotated.

Different from CNNs, ViTs
receive raw image patches as input and process   patch embeddings
via attention mechanism.
Due to the huge architectural differences,
the preliminary studies that simply
extended the transfer-based attacks  designed for CNNs to
ViTs  exhibited unsatisfactory  attacking effects~\cite{sia2023}.
To improve adversarial transferability,
later studies began to consider the unique features of ViTs
in the attack process.
ATA~\cite{ATA} generated transferable adversarial examples by  intensively activating  the uncertainty of patch-wise attention regions.
PNA~\cite{PNA} skipped the gradient of attention blocks during backpropagation for better  transferability.
TGR~\cite{TGR} generated adversarial samples by regularizing the back-propagated gradient in each internal
block of ViTs in a token-wise manner.
Unlike the above approaches attacking homogeneous models,
DeCoWA \cite{lin2024boosting} designed a deformation-constrained warping attack strategy to enhance the cross-architecture transferability.
MIG~\cite{ma2023transferable} used integrated gradients to replace regular gradients and utilized momentum iteration methods to improve the transferability across CNNs and ViTs.

\subsection{Adversarial Defenses}

To enhance the robustness of DNNs against adversarial examples,
a variety of defense methods have been explored. 
Among these,
adversarial training 
typically augmented the training set with adversarial examples to improve model resilience~\cite{madry2018towards}. 
TRADES~\cite{zhang2019theoretically} introduced a regularization term to trade off accuracy and robustness.
Ensemble adversarial training~\cite{tramer2018ensemble} 
decoupled the adversarial example generation from the target model, thereby enhancing robustness against black-box attacks.
Instead of modifying the target models,
input transformation-based defenses attempted to remove adversarial perturbations   by preprocessing   input images. 
FD~\cite{liu2019feature} 
defended against adversarial examples by filtering perturbations by frequency-domain JPEG compression.
Random resizing and padding (R\&P)~\cite{RP} introduced stochastic geometric transformations to break the structure of adversarial noise.
HGD~\cite{HGD} employed a U-Net-based decoder to reconstruct semantic features under attack.
NRP~\cite{NRP} used a self-supervised mechanism to purify perturbed features and restore clean images.
DiffPure~\cite{diffpure} utilized diffusion models to purify adversarial examples, achieving remarkable effectiveness.
Apart from the approaches mentioned above, certified defenses provided provable robustness guarantees under bounded perturbations,
such as randomized smoothing (RS)~\cite{cohen2019certified}, interval bound propagation (IBP)~\cite{IBP2019} and CROWN-IBP~\cite{crownibp2020}.
RS transformed a base classifier into a robust one by averaging predictions over Gaussian-noised inputs, providing formal robustness bounds.
IBP directly estimated worst-case logits under bounded input variation,
and CROWN-IBP~\cite{crownibp2020} improved the tightness of such bounds by advanced linear relaxations.
In this paper, we adopt the advanced defenses to assess the robustness of  adversarial examples.

\begin{algorithm}[!t]
	\caption{The \textsf{NAMEA} algorithm} \label{alg}
	\begin{algorithmic}[1]
		\REQUIRE{Surrogate models $\{f_{1},\ldots,f_{N}\}$, input sample $(x,y)$,
			perturbation budget $\epsilon$, step size $\alpha$, number of outer iterations $T$,
			number of inner loops   $K$, random sequence of model indices $S$.
		}
		\ENSURE {Adversarial example $x_{adv}$}.  \\
		\STATE  $x_{adv}^0=x$
		\FOR{$ t =0$ to $T-1$}
		\STATE $x^0_{tr} = x^t_{adv}$; $x^0_{te} = x^t_{adv}$
		\FOR{$k=0$ to $K-1$}
		\STATE Select a surrogate model via model indices $S$
		\STATE \textcolor{gray}{\# Meta-training step}
		\STATE{Calculate meta-training gradient $g^{k+1}_{tr}$ and adversarial example $x^{k+1}_{tr}$ with Eq.~(5)-Eq.~(6)}
		\STATE \textcolor{gray}{\# Meta-testing step}
		\STATE{Calculate the attention mask $\mathbb{M}_k$  with Eq.~(7) }
		\STATE{   Calculate the non-attention mask as $\overline{\mathbb{M}}_k=\mathbf{1}-\mathbb{M}_k$ }
		\STATE{Mask the attention area of ${x}^{k}_{te}$ with Eq.~(8)}
		\STATE{Calculate meta-testing gradient $g^{k+1}_{te}$ with Eq.~(9)}
		\STATE{Optimize meta-testing gradient $g^{k+1}_{te}$ with Eq.~(12)-Eq.~(14)}
		\STATE{Calculate adversarial example ${x}^{k+1}_{te}$ with Eq.~(10)}
		\ENDFOR
		\STATE \textcolor{gray}{\# Final update step}
		\STATE{Update final gradient $g^{t+1}$ with Eq.~(11)}
		\STATE Update final adversarial sample $x^{t+1}_{adv}$ with Eq.~(2)
		\ENDFOR
		\RETURN $x_{adv}$ = $x^T_{adv}$
	\end{algorithmic}
\end{algorithm}

\section{Algorithm Description of \textsf{NAMEA}}
\label{sec:alg}
As a plug-and-play method,
\textsf{NAMEA} can be combined with various gradient-based attacks, such as I-FGSM and MI-FGSM. 
To clearly illustrate the working process, we combine   \textsf{NAMEA} with I-FGSM
and present  the overall procedure    in Alg.~\ref{alg}.
Besides,  we should
ensure that each surrogate model is selected at least once in every $N$ consecutive inner iterations.
Let $S$ denote
a random sequence of model indices, which will be reshuffled after every $N$ iterations.
In implementation, we choose the $k'$-th element of $S$ as
the model index  for  the $k$-th inner iteration, 
where $k'=(k \mod N)+1$.

\begin{table}[!htp]
	\scalebox{0.75}{
		\begin{tabular}{ccccccccccc}
			\toprule
			\multirow{2}{*}{Attack}&\multicolumn{5}{c}{ViTs}&\multicolumn{5}{c}{CNNs}  \\
			\cmidrule(lr){2-6} \cmidrule(lr){7-11}
			&ViT-B&DeiT-B&Swin-B&Swin-S&\cellcolor[gray]{0.9}\textbf{Avg.}&RN50&WRN101&BiT50&BiT101&\cellcolor[gray]{0.9}\textbf{Avg.} \\ \hline
			Ens&7.1&16.0&15.5&22.7&\cellcolor[gray]{0.9}15.3&35.6&17.9&12.1&10.1&\cellcolor[gray]{0.9}18.9  \\
			SVRE&14.4&37.6&37.1&51.7&\cellcolor[gray]{0.9}35.2&64.0&31.9&21.3&17.9&\cellcolor[gray]{0.9}33.8  \\
			AdaEA&26.3&50.5&40.1&54.5&\cellcolor[gray]{0.9}42.9&57.6&33.7&28.7&24.7&\cellcolor[gray]{0.9}36.2 \\
			CWA&56.4&73.3&52.2&62.6&\cellcolor[gray]{0.9}61.1&47.2&34.9&44.1&41.4&\cellcolor[gray]{0.9}41.9  \\
			SMER&39.5&70.0&63.2&75.3&\cellcolor[gray]{0.9}62.0&75.7&45.9&34.1&30.2&\cellcolor[gray]{0.9}46.5  \\
			CSA&36.8&58.9&50.8&77.8&\cellcolor[gray]{0.9}56.1&73.2&40.6&36.7&32.5&\cellcolor[gray]{0.9}45.8  \\
			\textbf{Ours}&\textbf{77.4}&\textbf{94.4}&\textbf{84.6}&\textbf{92.7}&\cellcolor[gray]{0.9}\textbf{87.3}&\textbf{87.5}&\textbf{64.1}&\textbf{57.6}&\textbf{53.2}&\cellcolor[gray]{0.9}\textbf{65.6}  \\ \hline
			Ens&31.9&45.7&47.1&56.3&\cellcolor[gray]{0.9}45.3&75.0&51.3&40.8&37.1&\cellcolor[gray]{0.9}51.1  \\
			SVRE&59.7&75.1&75.2&82.7&\cellcolor[gray]{0.9}73.2&89.4&72.7&63.4&60.5&\cellcolor[gray]{0.9}71.5  \\
			AdaEA&61.3&79.3&72.4&81.3&\cellcolor[gray]{0.9}73.6&82.0&61.1&57.4&52.7&\cellcolor[gray]{0.9}63.3 \\
			CWA&77.8&93.9&90.7&95.4&\cellcolor[gray]{0.9}89.5&92.7&78.5&74.6&72.2&\cellcolor[gray]{0.9}79.5  \\
			SMER&86.7&93.5&89.9&93.2&\cellcolor[gray]{0.9}90.8&94.1&88.6&82.3&78.9&\cellcolor[gray]{0.9}86.0  \\
			CSA&85.2&91.5&87.5&91.2 &\cellcolor[gray]{0.9}88.9&92.0&85.5&77.6&73.2&\cellcolor[gray]{0.9}82.1  \\
			\textbf{Ours}&\textbf{91.9}&\textbf{97.5}&\textbf{96.7}&\textbf{98.3}&\cellcolor[gray]{0.9}\textbf{96.1}&\textbf{97.6}&\textbf{91.8}&\textbf{87.6}&\textbf{85.4}&\cellcolor[gray]{0.9}\textbf{90.6}  \\
			\bottomrule
		\end{tabular}
	}
	\centering
	\caption{Comparison of   ASRs ($\%$) between \textsf{NAMEA} and  baselines on CIFAR-10. The first and second rows are the I-FGSM and DI-MI-FGSM attacks, respectively. For all the tables, the best results are highlighted in bold.  }  \label{attack-cifar10}
\end{table}

\begin{table}[!t]
	\scalebox{0.75}{
		\begin{tabular}{ccccccccccc}
			\toprule
			\multirow{2}{*}{Attack}&\multicolumn{5}{c}{ViTs}&\multicolumn{5}{c}{CNNs}  \\
			\cmidrule(lr){2-6} \cmidrule(lr){7-11}
			&ViT-B&DeiT-B&Swin-B&Swin-S&\cellcolor[gray]{0.9}\textbf{Avg.}&RN50&WRN101&BiT50&BiT101&\cellcolor[gray]{0.9}\textbf{Avg.} \\ \hline
			Ens&34.1&58.1&58.6&70.0&\cellcolor[gray]{0.9}55.2&67.3&51.3&46.3&34.2&\cellcolor[gray]{0.9}49.8  \\
			SVRE&33.0&54.6&61.3&70.3&\cellcolor[gray]{0.9}54.8&68.4&53.5&49.6&37.2&\cellcolor[gray]{0.9}52.2  \\
			AdaEA&44.4&64.5&59.7&71.2&\cellcolor[gray]{0.9}60.0&73.5&61.9&56.2&44.8&\cellcolor[gray]{0.9}59.1 \\
			CWA&61.0&75.1&59.9&73.7&\cellcolor[gray]{0.9}67.4&62.7&62.0&55.0&49.8&\cellcolor[gray]{0.9}57.4  \\
			SMER&52.0&73.3&70.0&78.4&\cellcolor[gray]{0.9}68.4&73.9&60.6&56.8&44.6&\cellcolor[gray]{0.9}59.0  \\
			CSA&51.8&75.4&71.2&77.6&\cellcolor[gray]{0.9}69.0&74.2&62.5&58.9&46.7&\cellcolor[gray]{0.9}60.6  \\
			\textbf{Ours}&\textbf{75.7}&\textbf{88.9}&\textbf{84.8}&\textbf{89.5}&\cellcolor[gray]{0.9}\textbf{84.7}&\textbf{82.1}&\textbf{71.5}&\textbf{68.1}&\textbf{55.9}&\cellcolor[gray]{0.9}\textbf{69.4}  \\ \hline
			Ens&71.2&81.5&78.3&82.4&\cellcolor[gray]{0.9}78.4&82.5&74.5&71.8&62.3&\cellcolor[gray]{0.9}72.8  \\
			SVRE&73.0&83.2&82.6&87.0&\cellcolor[gray]{0.9}81.5&86.0&78.3&76.3&67.6&\cellcolor[gray]{0.9}77.1  \\
			AdaEA&74.0&85.3&80.5&86.3&\cellcolor[gray]{0.9}81.5&86.1&78.6&74.9&65.6&\cellcolor[gray]{0.9}76.3 \\
			CWA&77.5&88.3&83.8&87.1&\cellcolor[gray]{0.9}84.2&89.0&82.9&80.6&72.0&\cellcolor[gray]{0.9}81.1  \\
			SMER&81.8&88.9&85.6&88.7&\cellcolor[gray]{0.9}86.3&88.7&81.6&80.0&71.3&\cellcolor[gray]{0.9}80.4  \\
			CSA&79.8&86.5&83.8&86.1&\cellcolor[gray]{0.9}84.1&85.4&79.2&77.1&69.1&\cellcolor[gray]{0.9}77.7  \\
			\textbf{Ours}&\textbf{87.8}&\textbf{92.5}&\textbf{91.5}&\textbf{92.6}&\cellcolor[gray]{0.9}\textbf{91.1}&\textbf{91.7}&\textbf{96.5}&\textbf{84.3}&\textbf{77.5}&\cellcolor[gray]{0.9}\textbf{87.5}  \\
			\bottomrule
		\end{tabular}
	}
	\centering
	\caption{Comparison of   ASRs ($\%$) between \textsf{NAMEA} and  baselines on CIFAR-100. The first and second rows are the I-FGSM and DI-MI-FGSM attacks, respectively.}  \label{attack-cifar100}
\end{table}
\begin{table}[!t]
	\begin{tabular}{ccc}
		\toprule
		{Attack}
		&Per adv. example& GPU memory \\ \hline
		Ens&0.03 s&~~4.5 GB\\
		SVRE&0.21 s&10.2 GB  \\
		AdaEA&0.10 s&22.1 GB  \\
		CWA&0.05 s&~~4.5 GB \\
		SMER&0.70 s&~~6.1 GB  \\
		CSA&(69.6 h $\times$ 4 +) 0.51 s&22.3 GB \\
		{Ours}&0.91 s&~~6.5 GB  \\
		\bottomrule
	\end{tabular}
	\centering
	\caption{Computational overhead and GPU memory usage.}  \label{computation-cost}
\end{table}

\begin{figure*}[!t]
	\centering
	\includegraphics[scale=0.28]{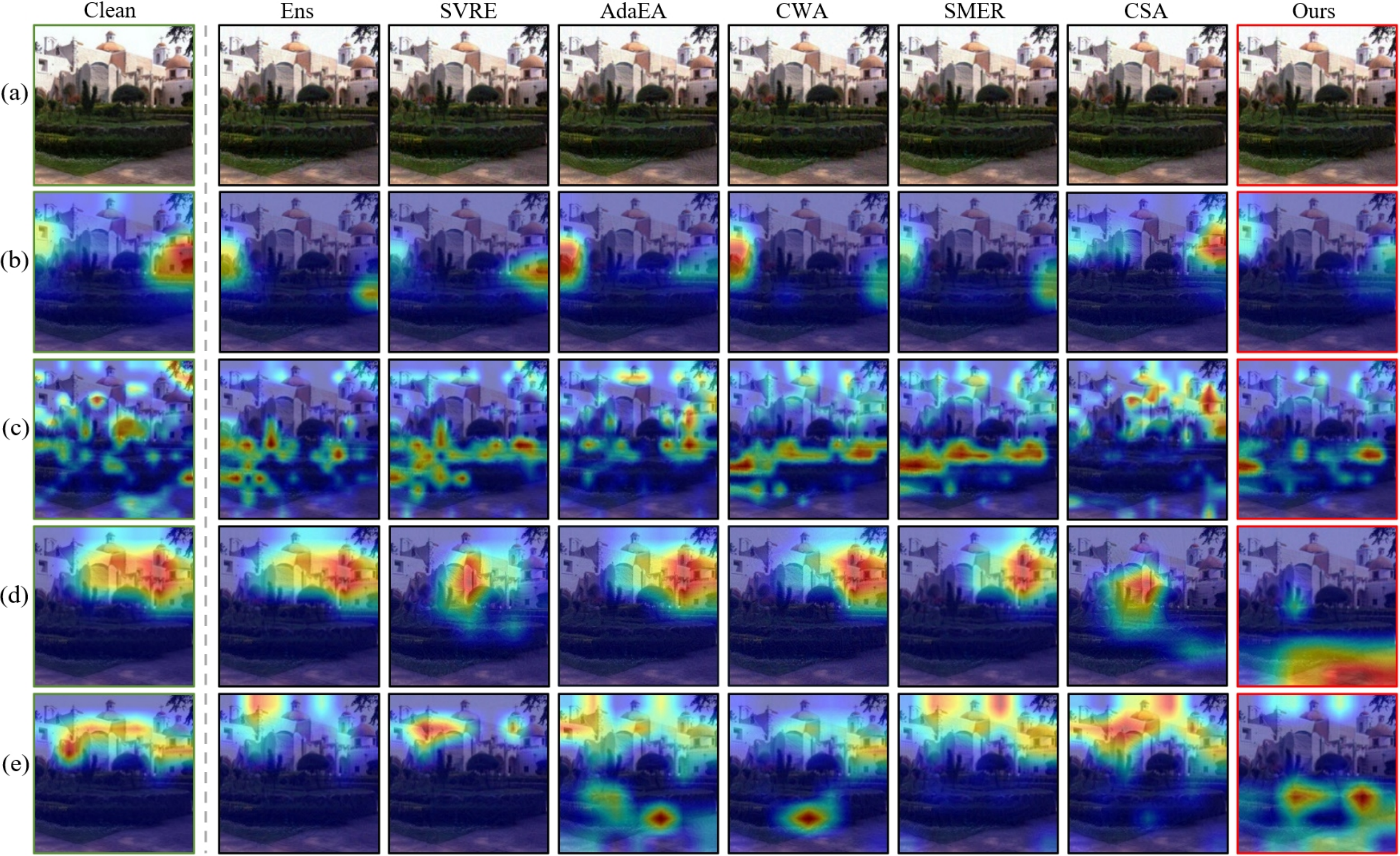}
	\caption{The   attention heatmaps of different input images.
		(a) shows input images including the clean image and the adversarial examples crafted by baselines and \textsf{NAMEA}.
		(b)(c) show the heatmaps on the surrogate models  RN18 and  ViT-T,  respectively.
		(d)(e)  show the heatmaps on target models WRN101 and Swin-T,  respectively. Base attack DI-MI-FGSM and   $\epsilon = 8/255$.} \label{att_heatmaps}
\end{figure*}
\section{Evaluation on CIFAR-10 and CIFAR-100}
\label{sec:experiments setup}

\textbf{Datasets and Models.} CIFAR-10~\cite{cifar} consists of 60,000  images of 10 categories, and
CIFAR-100~\cite{cifar} consists of 60,000  images of 100 categories.
We use the standard test sets in CIFAR-10 and CIFAR-100 to
evaluation the performance of \textsf{NAMEA} and baselines.
The test images are evenly distributed across all classes, with 1,000 (resp. 100) images per class for  CIFAR-10 (resp. CIFAR-100).
Following~~\cite{AdaEa},
we employ ViT-T\cite{vitb}, DeiT-T~\cite{deit}, ResNet-18 (RN18)~\cite{resenet}, and Inception-v3 (Inc-v3)~\cite{incv3} as the surrogate models.
The  target models  
are adopted from:
\textit{{ViT models:}} ViT-B~\cite{vitb}, DeiT-B~\cite{deit}, Swin-B~\cite{swin}
and Swin-S~\cite{swin}.
\textit{{CNN models:}} ResNet-50 (RN50)~\cite{resenet}, WideResNet-101 (WRN101), BiT-M-R50$\times$1 (BiT50) and BiT-M-R101 (BiT101)~\cite{widenet,kolesnikov2020big}.

\textbf{Baselines and Parameters.}
We compare the  ASR
with    six ensemble attacks:  Ens~\cite{ens}, SVRE~\cite{Svre}, AdaEA~\cite{AdaEa},  CWA~\cite{CWA}, SMER~\cite{SMER},
and CSA~\cite{CSA}, under the same ensemble settings and perturbation budget  $\epsilon =8/255$.
For the baselines and our \textsf{NAMEA},
we use I-FGSM  and DI-MI-FGSM    as the basic attacks.
The  hyper-parameters in the baselines follow the optimal setting in  the respective literature.
For a fair comparison, CSA  employs  7 checking points from each surrogate model, expanding  the ensemble scale to 28  models.
We set the number of outer iterations as $T = 10$ and
the number of internal loops as $K=16$,   using
step size   $\alpha = 0.8/255$ and the momentum decay    $\mu = 1.0$.
We use the same way as the main text to extract attention areas
under the attention  threshold    $\eta = 0.6$.

\textbf{Cross-Architecture Transferability.}
From   Table~\ref{attack-cifar10} and Table~\ref{attack-cifar100}, we can see that our \textsf{NAMEA} achieves superior adversarial transferability,
always performing  best when combining with different base attacks.
And we also  have the similar
observation  as in
the main text, i.e.,
\textsf{NAMEA} works  better under base attack DI-MI-FGSM  compared with base attack  I-FGSM.
The experiment results on CIFAR-10 and CIFAR-100
further confirmed the conclusions drawn in the main text, i.e.,
the non-attention areas of surrogate models contribute to improve cross-architecture transferability.

\begin{table*}[!t]
	\scalebox{0.78}{
		\begin{tabular}{ccccccccccccccccccccc}
			\toprule
			\multirow{2}{*}{Base}&\multirow{2}{*}{Attack}&\multicolumn{10}{c}{ViTs}&\multicolumn{9}{c}{CNNs}  \\
			\cmidrule(lr){3-12} \cmidrule(lr){13-21}
			\multirow{2}{*}{}&&ViT-B&PiT-B&CaiT-S&ViS&DeiT-B&TNT-S&LeViT&ConV&Swin-B&\cellcolor[gray]{0.9}\textbf{Avg.}&RN50&RN152&DN201&DN169&VGG16&VGG19&WRN101&BiT50&\cellcolor[gray]{0.9}\textbf{Avg.} \\ \hline
			\multirow{3}{*}{}&Ens&16.3&13.9&22.1&19.1&21.9&28.7&22.0&24.1&12.4&\cellcolor[gray]{0.9}20.1&26.4&15.1&33.4&34.9&41.7&40.5&27.1&32.7&\cellcolor[gray]{0.9}31.5 \\
			\multirow{3}{*}{FGSM}&SVRE&26.4&16.0&40.8&26.6&42.8&48.9&35.5&45.1&14.6&\cellcolor[gray]{0.9}33.0&40.2&22.0&47.5&50.1&58.6&56.7&35.9&45.6&\cellcolor[gray]{0.9}44.6  \\
			\multirow{3}{*}{}&AdaEA&21.9&15.9&27.8&22.1&28.5&32.8&23.7&30.9&15.2&\cellcolor[gray]{0.9}24.3&27.1&15.7&33.3&35.7&41.8&40.1&27.7&33.2&\cellcolor[gray]{0.9}31.8  \\
			\multirow{3}{*}{}&CWA&19.1&6.5&29.7&10.8&34.9&36.7&15.4&36.4&7.2&\cellcolor[gray]{0.9}21.9&10.8&5.0&17.0&19.0&33.3&33.0&12.7&21.0&\cellcolor[gray]{0.9}19.0  \\
			\multirow{3}{*}{}&SMER&10.8&6.0&16.4&8.7&18.1&24.7&12.8&22.1&6.5&\cellcolor[gray]{0.9}14.0&14.0&7.6&19.0&22.1&32.7&32.6&13.8&22.4&\cellcolor[gray]{0.9}20.5  \\
			\multirow{3}{*}{}&CSA&25.4&18.8&32.2&25.7&33.2&43.1&27.7&35.5&18.5&\cellcolor[gray]{0.9}28.9&30.6&19.4&38.6&41.7&46.3&44.6&32.0&39.0&\cellcolor[gray]{0.9}36.5  \\
			\multirow{3}{*}{}&\textbf{Ours}&\textbf{34.2}&\textbf{21.3}&\textbf{52.7}&\textbf{32.6}&\textbf{51.9}&\textbf{57.7}&\textbf{43.7}&\textbf{55.0}&\textbf{19.8}&\cellcolor[gray]{0.9}\textbf{41.0} &\textbf{47.2}&\textbf{27.2}&\textbf{54.5}&\textbf{57.1}&\textbf{66.9}&\textbf{64.1}&\textbf{43.3}&\textbf{51.5}&\cellcolor[gray]{0.9}\textbf{51.5}  \\ \hline
			\multirow{3}{*}{}&Ens&36.4&26.9&50.3&36.3&53.8&55.5&39.4&53.5&19.5&\cellcolor[gray]{0.9}41.3&43.2&28.7&56.7&57.7&56.7&55.8&43.6&50.8&\cellcolor[gray]{0.9}49.2  \\
			\multirow{3}{*}{TIM}&SVRE&26.0&20.6&38.5&31.0&39.5&49.6&32.8&41.0&15.0&\cellcolor[gray]{0.9}32.7&38.6&24.7&54.6&56.1&57.3&54.0&40.4&50.9&\cellcolor[gray]{0.9}47.1  \\
			\multirow{3}{*}{}&AdaEA&19.4&14.8&30.3&25.9&34.2&45.9&24.3&37.6&10.8&\cellcolor[gray]{0.9}24.0&28.0&16.8&50.1&49.2&51.8&50.1&34.1&48.7&\cellcolor[gray]{0.9}36.1  \\
			\multirow{3}{*}{}&CWA&32.3&19.1&49.1&30.3&52.2&61.2&38.0&56.8&14.5&\cellcolor[gray]{0.9}39.3&37.5&23.4&53.7&54.9&63.0&58.4&39.8&50.7&\cellcolor[gray]{0.9}47.8  \\
			\multirow{3}{*}{}&SMER&38.1&24.2&53.7&38.5&57.8&63.6&41.2&61.9&18.1&\cellcolor[gray]{0.9}44.1&41.5&29.1&58.9&59.6&64.0&60.0&45.4&55.6&\cellcolor[gray]{0.9}51.8  \\
			\multirow{3}{*}{}&CSA&44.6&28.8&59.6&43.2&61.9&64.8&46.3&61.2&19.6&\cellcolor[gray]{0.9}47.8&47.2&32.0&62.9&64.3&63.2&62.0&50.4&57.8&\cellcolor[gray]{0.9}55.0  \\
			\multirow{3}{*}{}&\textbf{Ours}&\textbf{50.1}&\textbf{34.5}&\textbf{69.0}&\textbf{49.0}&\textbf{71.6}&\textbf{74.3}&\textbf{55.0}&\textbf{71.1}&\textbf{25.9}&\cellcolor[gray]{0.9}\textbf{55.6}&\textbf{53.1}&\textbf{35.5}&\textbf{69.4}&\textbf{69.0}&\textbf{72.1}&\textbf{70.2}&\textbf{57.9}&\textbf{64.4}&\cellcolor[gray]{0.9}\textbf{61.5}  \\ \hline
			\multirow{3}{*}{ }&Ens&43.1&37.3&57.0&49.6&56.8&64.0&52.2&58.0&26.3&\cellcolor[gray]{0.9}49.4&43.2&28.7&56.7&57.7&56.7&55.8&43.6&50.8&\cellcolor[gray]{0.9}49.2  \\
			\multirow{3}{*}{DI-TIM-FGSM}&SVRE&35.5&31.1&52.0&45.2&51.8&66.5&48.9&54.5&20.2&\cellcolor[gray]{0.9}45.1&38.6&24.7&54.6&56.1&57.3&54.0&40.4&50.9&\cellcolor[gray]{0.9}47.1  \\
			\multirow{3}{*}{}&AdaEA&50.1&37.8&67.3&52.4&67.6&72.7&56.7&67.8&28.9&\cellcolor[gray]{0.9}55.7&28.0&16.8&50.1&49.2&51.8&50.1&34.1&48.7&\cellcolor[gray]{0.9}36.1  \\
			\multirow{3}{*}{}&CWA&47.9&39.8&70.3&55.9&69.9&79.0&63.4&70.8&27.3&\cellcolor[gray]{0.9}58.3&37.5&23.4&53.7&54.9&63.0&58.4&39.8&50.7&\cellcolor[gray]{0.9}47.8 \\
			\multirow{3}{*}{}&SMER&58.0&51.3&76.4&64.9&77.6&83.6&69.1&77.9&34.5&\cellcolor[gray]{0.9}65.9&41.5&29.1&58.9&59.6&64.0&60.0&45.4&55.6&\cellcolor[gray]{0.9}51.8 \\
			\multirow{3}{*}{}&CSA&53.5&46.4&68.0&58.9&68.6&73.5&61.7&68.1&34.7&\cellcolor[gray]{0.9}59.3&60.5&46.2&75.5&75.2&74.9&73.0&65.9&71.8&\cellcolor[gray]{0.9}67.9 \\
			\multirow{3}{*}{}&\textbf{Ours}&\textbf{65.9}&\textbf{56.8}&\textbf{80.4}&\textbf{71.6}&\textbf{82.2}&\textbf{87.3}&\textbf{73.9}&\textbf{81.7}&\textbf{39.6}&\cellcolor[gray]{0.9}\textbf{71.0}&\textbf{69.3}&\textbf{55.7}&\textbf{85.9}&\textbf{86.8}&\textbf{86.2}&\textbf{83.4}&\textbf{75.3}&\textbf{83.6}&\cellcolor[gray]{0.9}\textbf{78.3}
			\\
			\bottomrule
	\end{tabular} }
	\centering
	\caption{Comparison of   ASRs ($\%$) between \textsf{NAMEA} and  baselines under additional base attacks.  }  \label{attack-vit}
\end{table*}

\section{Comparison  of Overheads}\label{sec:computation}
All experiments are implemented by PyTorch, running on a server with NVIDIA GeForce RTX 4090 GPU.
We present the comparison of computational overheads and GPU memories in Table~\ref{computation-cost}.
The base attack is DI-MI-FGSM and the batch size is 20.
From the table, we can see that our \textsf{NAMEA}
can generate an adversarial example within one second, which is comparable to the SOTA baseline SMER.
Besides, the GPU memory usage of \textsf{NAMEA} is significantly lower than that of SVRE, AdaEA, and CSA.
For  AdaEA, the excessive   memory usage is attributed to saving large amounts of gradient  information for evaluating the adversarial ratio.
Notably, CSA   introduces extra  computational overhead (about 69.6 hours $\times$ 4) to train four surrogate models,
while consuming huge  GPU memory (about 22.3 GB) to save 28 checkpoints.
Specifically,   training single surrogate model typically requires 100 epochs, where each epoch needs to be   run  for  approximately 20 minutes.
Besides,   each epoch  requires  approximately  21.8 minutes in addition  to compute the accuracy gap between the training and validation sets.

\section{Visualization of  Attack Performance}
\label{sec:visualization}
To intuitively show the attack performance,
we visualize
the clean image and the adversarial examples generated by \textsf{NAMEA} and   baselines,
and employ Grad-CAM~\cite{CAM} to generate corresponding  attention heatmaps  under multiple surrogate and target models.
As shown in Figs.~\ref{att_heatmaps}(b)(c),
the attention of   surrogate models changes greatly  on all
the generated adversarial images compared with the clean
image,  indicating that all these  adversarial examples can effectively fool surrogate models. 
However,
Figs.~\ref{att_heatmaps}(d)(e) reveal that
compared to baselines, the adversarial examples generated by our \textsf{NAMEA} can
further shift the models' focus away from the original areas  in the black-box setting.
This demonstrates that \textsf{NAMEA}  achieves the highest
cross-architecture   transferability among competitors.

\section{Supplementary Experimental Results on Cross-Architecture Transferability}
\label{sec:supplementary results}

\begin{table}[!t]
	\scalebox{0.8}{
		\begin{tabular}{cccccccc}
			\toprule
			{Attack}&Twins-P-S&Twins-P-B&Twins-P-L&CoaT-Mini&CoaT-T&CoaT-S&\cellcolor[gray]{0.9}\textbf{Avg.} \\ \hline
			Ens&37.2&32.9&31.1&32.2&39.5&32.9&\cellcolor[gray]{0.9}34.3\\
			SVRE&37.8&32.8&28.9&34.4&42.3&34.7&\cellcolor[gray]{0.9}35.2  \\
			AdaEA&41.8&35.6&32.1&36.0&45.6&37.5&\cellcolor[gray]{0.9}38.1  \\
			CWA&31.3&25.8&23.2&27.5&37.4&25.6&\cellcolor[gray]{0.9}28.5  \\
			SMER&41.1&35.1&35.1&37.7&47.6&35.6&\cellcolor[gray]{0.9}38.7  \\
			CSA&44.4&40.4&37.8&39.2&50.3&40.4&\cellcolor[gray]{0.9}42.1  \\ \textbf{Ours}&\textbf{51.1}&\textbf{46.3}&\textbf{43.0}&\textbf{46.3}&\textbf{58.5}&\textbf{44.8}&\cellcolor[gray]{0.9}\textbf{48.3}  \\
			\hline
			Ens&50.2&45.8&42.7&42.0&52.6&42.7&\cellcolor[gray]{0.9}46.0\\
			SVRE&48.4&39.4&37.1&39.0&49.6&40.2&\cellcolor[gray]{0.9}42.3  \\
			AdaEA&54.2&49.0&45.5&45.8&58.1&48.0&\cellcolor[gray]{0.9}50.1  \\
			CWA&56.0&50.3&46.4&46.7&61.3&45.6&\cellcolor[gray]{0.9}51.1  \\
			SMER&66.0&59.2&54.8&54.9&67.3&56.5&\cellcolor[gray]{0.9}59.8  \\
			CSA&58.2&53.7&50.9&51.0&61.3&51.2&\cellcolor[gray]{0.9}54.4  \\ \textbf{Ours}&\textbf{72.3}&\textbf{65.9}&\textbf{61.7}&\textbf{62.2}&\textbf{72.7}&\textbf{61.7}&\cellcolor[gray]{0.9}\textbf{66.1} \\
			\hline
			Ens&51.6&46.7&44.7&46.7&56.0&46.9&\cellcolor[gray]{0.9}48.8\\
			SVRE&59.8&54.2&49.7&52.8&63.3&51.6&\cellcolor[gray]{0.9}55.2 \\
			AdaEA&52.5&47.4&44.5&47.5&58.6&48.7&\cellcolor[gray]{0.9}49.9  \\
			CWA&60.1&54.4&50.8&52.5&64.4&51.1&\cellcolor[gray]{0.9}55.6  \\
			SMER&72.6&66.7&63.9&65.0&74.1&64.4&\cellcolor[gray]{0.9}67.8  \\
			CSA&61.0&56.8&53.3&53.5&64.1&55.1&\cellcolor[gray]{0.9}57.3  \\ \textbf{Ours}&\textbf{78.9}&\textbf{72.2}&\textbf{72.6}&\textbf{73.1}&\textbf{81.2}&\textbf{70.9}&\cellcolor[gray]{0.9}\textbf{74.8} \\
			\bottomrule
	\end{tabular} }
	\centering
	\caption{Comparison of   ASRs ($\%$) against hybrid models.
		The first, second and third rows are the MI-FGSM, DI-TIM-FGSM and DI-MI-FGSM attacks, respectively.}  \label{tab:attack-hybrid-models}
\end{table}

\begin{table}[!t]
	\scalebox{0.85}{
		\begin{tabular}{ccccccccc}
			\toprule
			$\epsilon$& Base & Ens & SVRE & AdaEA & CWA & SMER & CSA & \textbf{Ours} \\
			\hline
			\multirow{2}{*} {12/255} & MI-FGSM & 55.8 & 58.8 & 62.1 & 56.9 & 68.6 & 68.8 & \textbf{76.9} \\
			& DI-MI-FGSM & 71.4 & 76.3 & 72.3 & 80.1 & 88.5 & 79.3 & \textbf{91.8} \\
			\hline
			\multirow{2}{*}{16/255} & MI-FGSM & 64.6 & 68.0 & 70.7 & 66.4 & 79.3 & 78.6 & \textbf{85.8} \\
			& DI-MI-FGSM & 78.7 & 84.3 & 79.1 & 87.4 & 91.1 & 85.6 & \textbf{96.2} \\
			\bottomrule
	\end{tabular} }	
	\centering
	\caption{Comparison of average ASRs (\%) on 9 ViTs   and 8 CNNs under different perturbation budgets.
	}  \label{tab:attack-budgets}
\end{table}

\begin{table}[!t]
	\scalebox{0.85}{
		\begin{tabular}{ccccccccc}
			\toprule
			Model No.& Base & Ens & SVRE & AdaEA & CWA & SMER & CSA & \textbf{Ours} \\
			\hline
			\multirow{2}{*} {6} & MI-FGSM & 53.7 & 57.7 & 60.3 & 57.9 & 66.7 & 68.3 & \textbf{75.2} \\
			& DI-MI-FGSM & 66.4 & 75.2 & 69.4 & 81.0 & 85.3 & 75.9 & \textbf{90.4} \\
			\hline
			\multirow{2}{*}{8} & MI-FGSM & 60.9 & 70.1 & 70.9 & 69.0 & 77.6 & 78.1 & \textbf{83.4} \\
			& DI-MI-FGSM & 72.0 & 81.7 & 74.1 & 86.8 & 90.1 & 88.2 & \textbf{95.2} \\
			\bottomrule
	\end{tabular} }	
	\centering
	\caption{Comparison of average ASRs (\%)  on 9 ViTs   and 8 CNNs under varying numbers of  surrogate models.
	}  \label{tab:more_surro}
\end{table}

\begin{table*}[!t]
	\scalebox{0.88}{
		\begin{tabular}{cccccccccccccccccc}
			\toprule
			\multirow{2}{*}{Attack}&\multicolumn{7}{c}{Defense Models} &\multicolumn{10}{c}{Defense Methods}  \\
			\cmidrule(lr){2-8} \cmidrule(lr){9-18}
			& Inc-v4 & IR-v2 & Inc-v3$_{adv}$ & Inc-v3$_{ens3}$ & Inc-v3$_{ens4}$ & IR-v2$_{ens}$ & \cellcolor[gray]{0.9}\textbf{Avg.} &R$\&$P&HGD & NIPS-r3& JPEG & RS   & NPR & FD  & Bit-RD & DiffPure & \cellcolor[gray]{0.9}\textbf{Avg.} \\
			\hline
			{Ens}&27.2   &44.4	& {64.1} &21.1	 & 26.5  & 40.6 &\cellcolor[gray]{0.9}37.3 &19.9  &7.4	& {15.7} &18.0	 & 10.2 & 10.3 & 18.4& 17.9 & 4.2 &\cellcolor[gray]{0.9}13.6	\\
			{SVRE}&31.2  &45.9	& {65.7} &21.8	 & 26.5  & 42.3 &\cellcolor[gray]{0.9}38.9 &22.0  &10.2	& {23.4} &19.6	 & 10.3 & 10.8 & 19.4& 19.5 & 3.7 &\cellcolor[gray]{0.9}15.4	\\
			{AdaEA}&36.5  &50.8	& {66.0} &24.8	 & 29.7  & 44.4 &\cellcolor[gray]{0.9}42.0 &31.9&14.0&27.2&29.9&12.3 &13.0 & 25.3&31.3 & 6.3 & \cellcolor[gray]{0.9}21.2 \\
			{CWA}&28.6	 &47.8	&66.1	 &23.9 	 &29.6	 & 45.3 &\cellcolor[gray]{0.9}40.2 &22.8&6.2&10.4&28.3 &13.9 &14.5 & 30.5&25.1& 8.1 & \cellcolor[gray]{0.9}17.8 \\
			{SMER}&38.1  & 51.5	& 66.8	 & 25.8  & 31.1  & 43.6 &\cellcolor[gray]{0.9}42.8 &29.1  &13.4	& 24.9	 & 28.0 & 11.6 & 12.6& 25.1& 28.5 &	5.3 &	\cellcolor[gray]{0.9}19.8	\\
			{CSA}&36.7  & 50.3	& 66.7	 & 26.0  & 30.1  & 45.0 &\cellcolor[gray]{0.9}42.5 &30.9  &14.3	& 28.3	 & 29.2 & 11.5 & 14.2& 25.3& 29.5 & 6.1 &	\cellcolor[gray]{0.9}21.0	\\
			{\textbf{Ours}}&\textbf{48.4}&\textbf{57.1}&\textbf{70.3}&\textbf{32.7} &\textbf{36.9} &\textbf{48.1} &\cellcolor[gray]{0.9}\textbf{48.9} &\textbf{40.2}&\textbf{19.4}&\textbf{34.4}&\textbf{39.8} &\textbf{14.2} &\textbf{17.0} &\textbf{33.7}&\textbf{40.3}& \textbf{10.7}&\cellcolor[gray]{0.9}\textbf{27.7} \\
			\hline
			Ens & 43.4 & 53.7 & 69.0 & 30.8 & 34.8 & 48.1 & \cellcolor[gray]{0.9}46.6 & 37.6 & 22.7 & 35.4 & 37.1 & 13.5 &14.8 & 35.2 & 38.2 & 14.3 & \cellcolor[gray]{0.9}27.6 \\
			SVRE & 48.0 & 55.6 & 70.4 & 33.5 & 35.5 & 49.0 & \cellcolor[gray]{0.9}48.7 &39.5  &25.1	& {43.1} &37.7	 & 13.9 & 15.2 & 35.6& 39.0 & 11.8 &\cellcolor[gray]{0.9}29.0 \\
			AdaEA & 46.3 & 54.7 & 68.9 & 31.9 & 35.0 & 48.0 & \cellcolor[gray]{0.9}47.5 &41.1&23.8&38.1&40.6&14.7 &15.1 &39.4&42.4 & 14.0 & \cellcolor[gray]{0.9}29.9  \\
			CWA & 50.7 & 56.5 & 70.6 & 29.8 & 35.4 & 48.2 & \cellcolor[gray]{0.9}48.5 &36.3&16.5&35.6&37.4 &13.8 &13.5 & 36.1&37.8& 10.7 & \cellcolor[gray]{0.9}26.4 \\
			SMER & 52.3 & 58.7 & 71.5 & 34.7 & 38.0& 51.3 & \cellcolor[gray]{0.9}51.1 &43.8  &25.3	& 42.0	 & 44.6 & 14.6 & 16.5& 41.9& 45.7 & 14.2	& \cellcolor[gray]{0.9}32.1 \\
			CSA & 51.1 & 58.9 & 71.5 & 37.5 & 40.1& 51.4 & \cellcolor[gray]{0.9}51.7 &47.5  &29.1	& 43.9	 & 47.3 & 14.9 & 17.6& 44.6& 49.1 & 14.9	& \cellcolor[gray]{0.9}34.3 \\
			\textbf{Ours} & \textbf{59.8} & \textbf{64.8} & \textbf{74.3} & \textbf{42.7} & \textbf{44.5} & \textbf{55.3} & \cellcolor[gray]{0.9}\textbf{56.9} &\textbf{53.3}&\textbf{34.8}&\textbf{49.2}&\textbf{54.7} &\textbf{17.4} &\textbf{21.1} &\textbf{51.1}&\textbf{55.6}& \textbf{20.0} & \cellcolor[gray]{0.9}\textbf{39.7} \\
			\bottomrule
	\end{tabular} }	
	\centering
	\caption{Comparison of ASRs (\%) against 6 defense models and 9 defense methods.
		The first and second rows are the I-FGSM and MI-FGSM attacks, respectively.
	}  \label{denfenseModels-IFGSM}
\end{table*}
\textbf{Additional Base Attacks.}
In the main text, we   assess the attack effects of \textsf{NAMEA} and baselines under three  base attacks:
I-FGSM, MI-FGSM, and DI-MI-FGSM.
To fully evaluate the performance, we  conduct experiments under three extra
base attacks:  FGSM~\cite{FGSM2014}, TIM and DI-TIM-FGSM~\cite{TIM2019}.
For   TIM and DI-TIM-FGSM, we  also set the momentum decay  as $\mu = 1.0$.
When used as base attacks of   \textsf{NAMEA} and baselines,
the sequence ranked by the    ascending   order of  attack    performance  is basically as follows:
FGSM, I-FGSM, TIM, MI-FGSM, DI-TIM-FGSM, and DI-MI-FGSM.
Besides, the results shown in Table~\ref{attack-vit}  are consistent with those in our main text,
thus demonstrating that \textsf{NAMEA} has superior cross-architecture transferability.

\textbf{Hybrid Target Models.}
In the main text, we  assess the attack effects of \textsf{NAMEA} and baselines on 9 ViTs and 8 CNNs.
To further demonstrate the cross-architecture transferability,
we conduct    evaluations on 6 hybrid architecture models,
including Twins-PCPVT-S (Twins-P-S), Twins-PCPVT-B (Twins-P-B),	Twins-PCPVT-L (Twins-P-L)~\cite{chu2021twins},	CoaT-Mini,	CoaT-T and	CoaT-S~\cite{xu2021co}.
From Table~\ref{tab:attack-hybrid-models}, we can observe that
our \textsf{NAMEA} and    baselines  basically achieve lower ASRs against hybrid target models,
but \textsf{NAMEA} consistently achieves superior performance across all evaluated models.

\textbf{Different Perturbation Budgets.}
In the main text, we evaluate the attack performance   under perturbation budget  $\epsilon = 8/255$.
For comprehensiveness, we    conduct experiments under   varying perturbation budgets.
Table~\ref{tab:attack-budgets} reports the average ASRs   on 9 ViTs   and 8 CNNs
under budgets $\epsilon =12/255$ and $\epsilon =16/255$. From this table, we can see that
as the  perturbation budget  increases, our \textsf{NAMEA} and    baselines achieve higher ASRs,
but \textsf{NAMEA} consistently outperforms the baselines across different budget levels.

\textbf{More Surrogate Models.}
In the main text, we evaluate the attack performance  under 4   surrogate models, i.e., ViT-T, DeiT-T, ResNet-18 (RN18), and Inception-v3 (Inc-v3).
To further validate the scalability, we evaluate the  performance under larger numbers of surrogate models:
(1) 6 surrogate models: ResNet-18, Inception-v3, BiT-M-R101 (BiT101), ViT-T, DeiT-T, and Swin-T;
(2) 8 surrogate models:  ResNet-18, Inception-v3, BiT-M-R101 (BiT101), DenseNet-121 (DN121), ViT-T, DeiT-T, Swin-T, and ConViT-B (ConV).
To align with the configuration
in the main text, 
CSA selects 4-5 checkpoints per model under the setting of  6 surrogate models  and
3-4 checkpoints per model under the setting of   8 surrogate models, ensuring that the   total number of surrogate models in different settings is fixed to 28.
Table~\ref{tab:more_surro} reports
the average ASRs on 9 ViTs and 8 CNNs under    varying numbers of  surrogate models,
where   the number of internal loops $K$ is set to be four times of the
surrogate model number.
From  Table~\ref{tab:more_surro}, we can observe that
as the  number of surrogate models   increases, our \textsf{NAMEA} and all  baselines achieve higher ASRs,
while  \textsf{NAMEA} consistently achieving the highest ASRs among all attacks.

\section{Supplementary Experimental Results on Robustness of Adversarial Examples}
Due to limited space, Table 2 and Table 3 of  the main text only show attack effects against 6 defense models and 9 defense methods under DI-MI-FGSM.
For completeness, we provide the results under I-FGSM and MI-FGSM in Table~\ref{denfenseModels-IFGSM}.
The results shown in  Table~\ref{denfenseModels-IFGSM}  are consistent with those in   the main text,
thus demonstrating that \textsf{NAMEA}  generates highly robust adversarial examples.

\section{Supplementary  Experimental Results on Ablation Studies}
\label{sec:more ablations}
Unless otherwise specified, the ablation experiments are assessed by the average ASRs against 9 ViTs   and 8 CNNs.

\begin{table*}[!t]
	\scalebox{0.78}{
		\begin{tabular}{ccccccccccccccccccccc}
			\toprule
			\multirow{2}{*}{Model}&\multirow{2}{*}{Attack}&\multicolumn{10}{c}{ViTs}&\multicolumn{9}{c}{CNNs}  \\
			\cmidrule(lr){3-12} \cmidrule(lr){13-21}
			\multirow{2}{*}{}&\multirow{2}{*}{}&ViT-B&Pit-B&CaiT-S&ViS&DeiT-B&TNT-S&LeVit&ConV&Swin-B&\cellcolor[gray]{0.9}\textbf{Avg.}&RN50&RN152&DN201&DN169&VGG16&VGG19&WRN101&BiT50&\cellcolor[gray]{0.9}\textbf{Avg.} \\ \hline
			{}&SVRE&6.3&4.4&7.7&5.4&9.7&14.5&6.6&13.0&4.1&\cellcolor[gray]{0.9}8.0&9.8&5.6&13.3&15.8&18.4&17.2&11.6&15.5&\cellcolor[gray]{0.9}13.4 \\
			{Inc-v3}&AdaEA&13.4&8.5&22.1&12.1&25.9&27.2&13.5&29.4&7.0&\cellcolor[gray]{0.9}17.7&12.7&7.7&20.5&23.2&28.6&26.1&15.3&21.8&\cellcolor[gray]{0.9}19.5 \\
			{DeiT-T}&CWA&15.2&5.8&25.2&9.0&31.6&29.6&9.9&34.1&6.4&\cellcolor[gray]{0.9}18.5&7.8&2.8&13.0&13.8&23.8&19.9&9.4&14.1&\cellcolor[gray]{0.9}13.1 \\
			{}&SMER&10.8&6.8&19.5&9.8&26.2&26.4&12.1&27.2&5.4&\cellcolor[gray]{0.9}16.0&12.3&6.2&16.5&19.8&26.0&23.0&13.8&19.5&\cellcolor[gray]{0.9}17.1 \\		{}&\textbf{Ours}&\textbf{15.7}&\textbf{9.7}&\textbf{27.8}&\textbf{12.6}&\textbf{32.3}&\textbf{31.9}&\textbf{15.4}&\textbf{34.9}&\textbf{7.3}&\cellcolor[gray]{0.9}\textbf{20.8}&\textbf{15.8}&\textbf{9.9}&\textbf{22.2}&\textbf{25.8}&\textbf{31.9}&\textbf{29.7}&\textbf{17.4}&\textbf{23.4}&\cellcolor[gray]{0.9}\textbf{22.0} \\ \hline			
			{}&SVRE&8.4&8.4&12.3&14.8&10.9&20.6&12.8&15.4&7.1&\cellcolor[gray]{0.9}12.3&26.2&13.6&31.3&36.5&38.9&35.7&23.8&28.8&\cellcolor[gray]{0.9}29.4 \\
			{RN18}&AdaEA&12.9&12.5&23.7&20.9&25.5&30.6&22.0&28.0&10.1&\cellcolor[gray]{0.9}20.7&32.9&18.3&40.6&43.7&48.1&46.3&29.1&33.1&\cellcolor[gray]{0.9}36.5  \\
			{Inc-v3}&CWA&17.4&10.2&32.1&14.0&36.9&38.5&16.8&37.7&9.7&\cellcolor[gray]{0.9}23.7&11.5&7.3&18.4&20.6&27.9&26.1&13.4&19.8&\cellcolor[gray]{0.9}18.1  \\
			{DeiT-T}&SMER&14.3&11.2&25.0&19.3&28.6&32.2&20.2&31.8&9.5&\cellcolor[gray]{0.9}21.3&29.9&15.6&36.1&39.3&47.2&43.5&26.2&34.1&\cellcolor[gray]{0.9}34.0 \\			{}&\textbf{Ours}&\textbf{18.4}&\textbf{16.3}&\textbf{34.9}&\textbf{25.8}&\textbf{37.1}&\textbf{41.6}&\textbf{30.0}&\textbf{38.8}&\textbf{12.5}&\cellcolor[gray]{0.9}\textbf{28.4}&\textbf{40.9}&\textbf{22.0}&\textbf{47.0}&\textbf{51.4}&\textbf{56.1}&\textbf{53.3}&\textbf{35.5}&\textbf{40.2}&\cellcolor[gray]{0.9}\textbf{43.3}  \\ \hline			
			{}&SVRE&14.3&9.8&26.0&17.7&25.9&29.8&17.2&28.7&9.4&\cellcolor[gray]{0.9}19.9&26.9&12.7&31.6&36.0&43.4&39.1&22.9&28.8&\cellcolor[gray]{0.9}30.2  \\
			{RN18}&AdaEA&21.5&13.6&35.3&23.8&39.1&37.4&24.0&39.9&13.0&\cellcolor[gray]{0.9}27.5&31.0&14.9&37.3&42.2&47.9&43.1&27.0&34.6&\cellcolor[gray]{0.9}34.8  \\
			{DeiT-T}&CWA&25.5&10.6&39.9&16.1&47.6&44.8&19.8&47.6&11.1&\cellcolor[gray]{0.9}29.2&12.1&6.5&20.4&22.5&34.5&32.3&14.5&25.2&\cellcolor[gray]{0.9}21.0  \\
			{ViT-T}&SMER&27.1&13.5&40.7&22.4&42.7&41.6&23.4&46.6&12.9&\cellcolor[gray]{0.9}30.1&29.9&14.8&37.3&41.4&48.0&44.3&25.3&32.6&\cellcolor[gray]{0.9}34.2 \\			{}&\textbf{Ours}&\textbf{33.7}&\textbf{18.3}&\textbf{51.4}&\textbf{28.6}&\textbf{54.4}&\textbf{50.5}&\textbf{30.8}&\textbf{56.6}&\textbf{16.4}&\cellcolor[gray]{0.9}\textbf{37.9}&\textbf{37.5}&\textbf{19.8}&\textbf{45.0}&\textbf{48.4}&\textbf{57.1}&\textbf{53.1}&\textbf{33.4}&\textbf{39.8}&\cellcolor[gray]{0.9}\textbf{41.8}  \\ \hline
			
			{RN18}&SVRE&13.1&11.5&21.9&19.2&23.2&28.2&19.3&23.9&10.1&\cellcolor[gray]{0.9}18.9&29.0&16.2&34.8&39.5&42.1&28.9&26.0&32.5&\cellcolor[gray]{0.9}32.4  \\
			{Inc-v3}&AdaEA&25.1&17.6&39.2&27.5&40.4&40.2&28.8&42.7&15.6&\cellcolor[gray]{0.9}30.8&38.7&21.1&47.0&50.1&53.0&48.4&34.5&39.6&\cellcolor[gray]{0.9}41.6  \\
			{ViT-T}&CWA&27.8&10.6&41.5&16.7&49.9&46.7&21.1&48.8&11.7&\cellcolor[gray]{0.9}30.5&12.9&6.9&20.8&22.6&34.3&32.1&15.2&25.5&\cellcolor[gray]{0.9}21.3  \\
			{DeiT-T}&SMER&27.4&16.4&42.6&26.0&43.9&44.7&27.7&48.9&15.4&\cellcolor[gray]{0.9}32.6&33.2&18.4&43.1&45.7&50.0&48.4&31.4&39.6&\cellcolor[gray]{0.9}38.7 \\		{}&\textbf{Ours}&\textbf{43.0}&\textbf{25.5}&\textbf{61.2}&\textbf{38.0}&\textbf{63.0}&\textbf{61.2}&\textbf{42.9}&\textbf{63.6}&\textbf{21.8}&\cellcolor[gray]{0.9}\textbf{46.7}&\textbf{46.2}&\textbf{26.4}&\textbf{55.8}&\textbf{58.5}&\textbf{64.4}&\textbf{60.7}&\textbf{43.8}&\textbf{52.1}&\cellcolor[gray]{0.9}\textbf{51.0}  \\ \hline
			
			{RN18}&SVRE&19.0&18.7&30.4&32.9&31.4&37.0&25.1&31.7&45.2&\cellcolor[gray]{0.9}30.2&33.7&16.2&35.9&40.5&48.2&45.3&25.5&31.5&\cellcolor[gray]{0.9}34.6  \\
			{ViT-T}&AdaEA&28.6&22.5&44.1&36.4&48.5&46.8&32.6&48.5&40.1&\cellcolor[gray]{0.9}38.7&39.1&21.4&43.5&49.8&54.6&51.9&32.9&39.9&\cellcolor[gray]{0.9}41.6  \\
			{DeiT-T}&CWA&16.4&22.2&25.7&31.4&26.4&37.4&24.2&24.8&49.3&\cellcolor[gray]{0.9}28.6&20.8&13.8&23.7&26.1&34.9&31.7&17.7&26.4&\cellcolor[gray]{0.9}24.4  \\
			{Swin-T}&SMER&32.9&23.0&48.2&34.4&50.5&50.1&32.6&52.3&45.2&\cellcolor[gray]{0.9}41.0&35.9&18.8&41.6&46.6&53.9&50.2&29.7&38.3&\cellcolor[gray]{0.9}39.4 \\			{}&\textbf{Ours}&\textbf{43.5}&\textbf{29.9}&\textbf{61.6}&\textbf{45.0}&\textbf{63.5}&\textbf{61.9}&\textbf{43.2}&\textbf{64.2}&\textbf{56.1}&\cellcolor[gray]{0.9}\textbf{52.1}&\textbf{46.1}&\textbf{25.7}&\textbf{51.9}&\textbf{55.7}&\textbf{62.8}&\textbf{58.7}&\textbf{38.5}&\textbf{47.6}&\cellcolor[gray]{0.9}\textbf{48.4}  \\ \hline
			
			{RN18}&SVRE&9.3&12.4&16.2&19.0&14.3&22.2&16.7&18.1&10.2&\cellcolor[gray]{0.9}{15.4}&30.4&18.6&37.8&40.4&41.5&38.6&28.1&43.9&\cellcolor[gray]{0.9}{34.9}  \\
			{Inc-v3}&AdaEA&17.6&15.8&28.2&26.8&28.6&34.1&26.3&35.3&13.6&\cellcolor[gray]{0.9}25.1&39.8&24.0&49.3&52.7&53.8&50.8&37.8&51.6&\cellcolor[gray]{0.9}45.0  \\
			{BiT-101}&CWA&23.1&13.7&38.2&23.2&40.9&41.5&21.7&44.6&12.6&\cellcolor[gray]{0.9}28.8&16.4&10.1&22.5&24.8&31.8&29.6&17.8&26.7&\cellcolor[gray]{0.9}22.5  \\
			{DeiT-T}&SMER&16.8&15.5&27.3&24.2&29.9&35.5&24.2&33.0&12.6&\cellcolor[gray]{0.9}24.3&35.0&21.7&45.1&47.5&50.3&46.6&33.2&49.4&\cellcolor[gray]{0.9}41.1 \\			{}&\textbf{Ours}&\textbf{24.0}&\textbf{21.2}&\textbf{41.0}&\textbf{34.2}&\textbf{41.6}&\textbf{47.3}&\textbf{35.6}&\textbf{45.5}&\textbf{18.7}&\cellcolor[gray]{0.9}\textbf{34.3}&\textbf{50.3}&\textbf{31.1}&\textbf{59.2}&\textbf{59.9}&\textbf{65.2}&\textbf{60.0}&\textbf{45.5}&\textbf{61.2}&\cellcolor[gray]{0.9}\textbf{54.1}  \\ \hline
			
			{}&SVRE&4.8&8.2&6.4&13.0&5.5&14.5&11.6&9.2&6.7&\cellcolor[gray]{0.9}8.9&24.1&14.6&29.0&33.1&34.1&32.1&23.1&34.4&\cellcolor[gray]{0.9}28.1 \\
			{RN18}&AdaEA&6.3&9.5&8.7&17.3&7.3&18.4&15.8&10.7&8.0&\cellcolor[gray]{0.9}11.3&31.5&17.9&39.2&43.8&43.0&42.1&28.5&43.1&\cellcolor[gray]{0.9}36.1 \\
			{Inc-v3}&CWA&3.9&4.5&3.0&7.2&4.0&11.4&7.5&4.8&5.6&\cellcolor[gray]{0.9}5.8&13.3&7.8&18.5&21.1&22.3&21.7&15.4&35.7&\cellcolor[gray]{0.9}19.5 \\
			{BiT-101}&SMER&5.5&9.1&7.7&14.8&5.7&18.1&12.5&7.3&7.7&\cellcolor[gray]{0.9}10.0&31.2&16.4&35.3&37.9&41.5&39.4&27.4&41.7&\cellcolor[gray]{0.9}33.9 \\		{}&\textbf{Ours}&\textbf{6.5}&\textbf{11.8}&\textbf{10.0}&\textbf{20.7}&\textbf{8.6}&\textbf{22.2}&\textbf{19.0}&\textbf{12.7}&\textbf{9.2}&\cellcolor[gray]{0.9}\textbf{13.4}&\textbf{37.9}&\textbf{21.1}&\textbf{44.7}&\textbf{47.7}&\textbf{50.9}&\textbf{47.9}&\textbf{34.1}&\textbf{50.1}&\cellcolor[gray]{0.9}\textbf{41.8}  \\ \hline
			
			{}&SVRE&18.1&12.6&28.5&19.7&30.7&31.7&14.7&32.5&44.5&\cellcolor[gray]{0.9}25.9&11.7&5.8&13.4&15.3&24.5&22.4&10.4&18.8&\cellcolor[gray]{0.9}15.3  \\
			{ViT-T}&AdaEA&30.0&19.2&42.1&29.2&49.6&42.2&24.8&47.3&47.6&\cellcolor[gray]{0.9}36.9&17.2&9.0&22.0&25.7&32.4&30.4&16.3&26.0&\cellcolor[gray]{0.9}22.4  \\
			{DeiT-T}&CWA&16.0&22.2&24.3&29.2&26.0&34.7&23.8&24.8&48.3&\cellcolor[gray]{0.9}27.7&20.5&13.5&23.4&25.1&33.7&32.1&16.4&25.4&\cellcolor[gray]{0.9}23.8  \\
			{Swin-T}&SMER&31.4&17.0&42.8&24.7&49.3&43.3&22.3&49.3&45.6&\cellcolor[gray]{0.9}36.2&16.0&8.4&19.6&23.4&34.3&31.0&14.5&24.9&\cellcolor[gray]{0.9}21.5 \\		{}&\textbf{Ours}&\textbf{34.4}&\textbf{19.4}&\textbf{50.7}&\textbf{27.5}&\textbf{54.7}&\textbf{51.1}&\textbf{28.0}&\textbf{54.3}&\textbf{49.7}&\cellcolor[gray]{0.9}\textbf{41.1}&\textbf{21.7}&\textbf{15.3}&\textbf{25.7}&\textbf{27.0}&\textbf{39.5}&\textbf{36.7}&\textbf{17.9}&\textbf{28.8}&\cellcolor[gray]{0.9}\textbf{26.6}  \\
			\bottomrule
	\end{tabular} }
	\centering
	\caption{Comparison of  ASRs ($\%$) between \textsf{NAMEA}  and baselines under different ensemble setting. Base attack    I-FGSM.}  \label{tab:attack-nums}
\end{table*}

\textbf{The Impact of Ensemble Settings.}
The ensemble settings in the main text employ four surrogate models:
ViT-T, DeiT-T, ResNet-18 (RN18), and Inception-v3 (Inc-v3).
As the number of CNNs and ViTs in the surrogate models has   noticeable effect on our method,
we investigate the performance under different  ensemble settings:  varying numbers of surrogate  models and varying ratios of CNNs to ViTs.
Note that, CSA leverages multiple checkpoints per model,
leading to a significantly larger   number of surrogate models than other methods.
Thus, it is unfair to make a direct comparison between CSA and other methods. 
So we exclude CSA and Ens (that always performs the worst)  from this ablation.
In the experiments, we introduce    new  models including
BiT-101~\cite{kolesnikov2020big} and Swin-T~\cite{swin},
and set the number of internal loops $K$ to be four times of the surrogate model number.

From Table~\ref{tab:attack-nums}, we can  see  that   \textsf{NAMEA} always performs best  under different ensemble settings.
Among them,    \textsf{NAMEA}  achieves the best performance when the ratio of CNNs to ViTs is 1:3.
We attribute this performance gain to the  comprehensive global information captured by
ViTs,  facilitating the extraction of transferable adversarial information.
Besides, we have the following observations:
\textcircled{1}  The overall performance of  \textsf{NAMEA}  and baselines  increase with the number of surrogate models.
This is because more surrogate models can capture more universal adversarial information.
\textcircled{2} When we keep the number of surrogate models   constant and retain at least one ViT/CNN,
increasing the proportion of ViTs will accordingly increase the ASRs on ViTs, but has  little impact on  CNNs.
\textcircled{3}  From the last two rows of Table~\ref{tab:attack-nums},
we observe that when the surrogate models only include    CNNs, the ASR on ViTs notably decreases. 
But when  only ViTs are used, there is no stark  difference between the ASRs on ViTs and CNNs.
These phenomena indicate that the
adversarial perturbations crafted from  ViTs are more transferable than those from CNNs.

\begin{figure*}[!t]
	\centering
	\begin{minipage}[c]{0.45\textwidth}
		\centering
		\includegraphics[width=\linewidth]{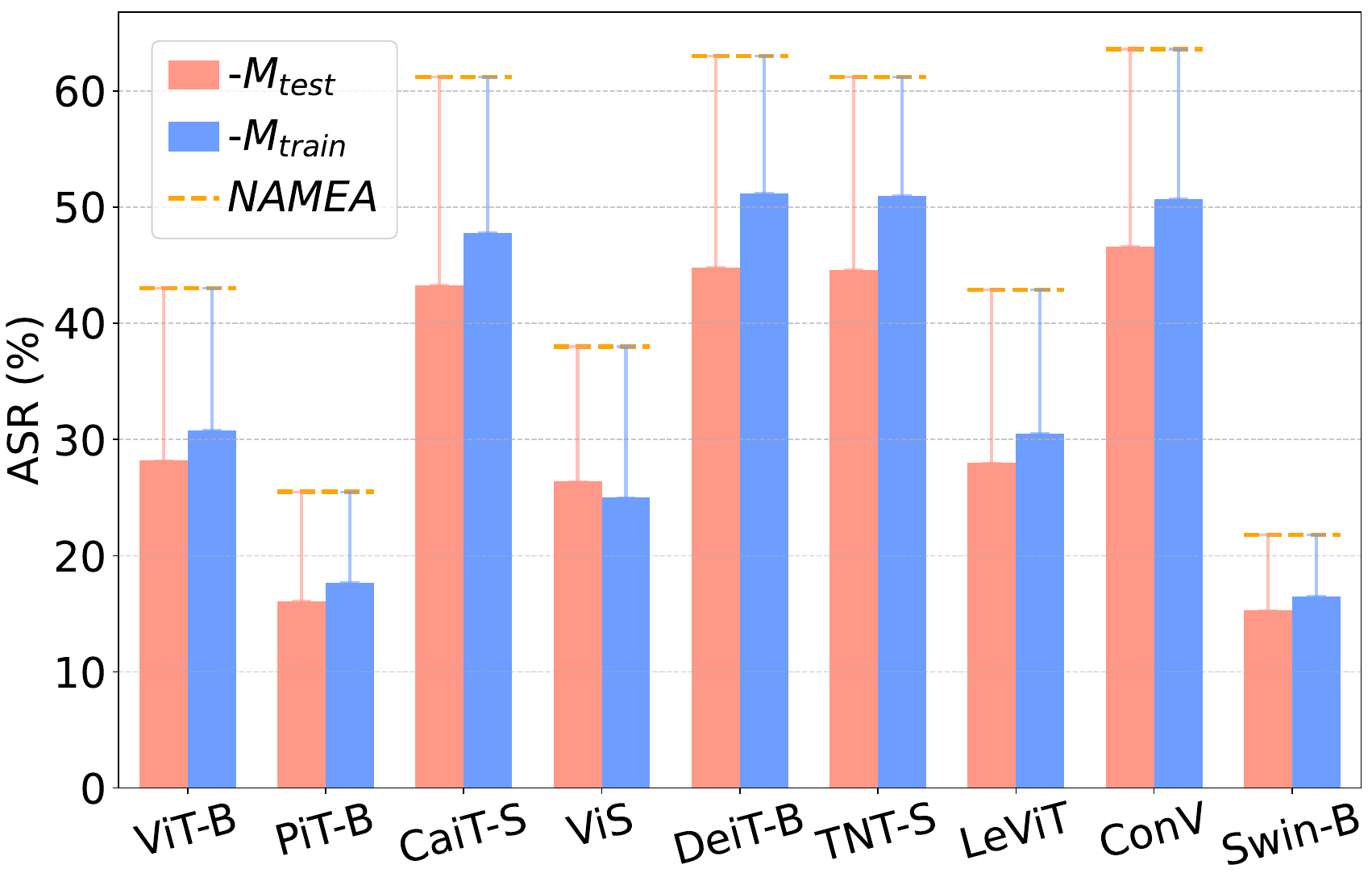}
	\end{minipage}
	\hspace{1.2em}
	\begin{minipage}[c]{0.45\textwidth}
		\centering
		\includegraphics[width=\linewidth]{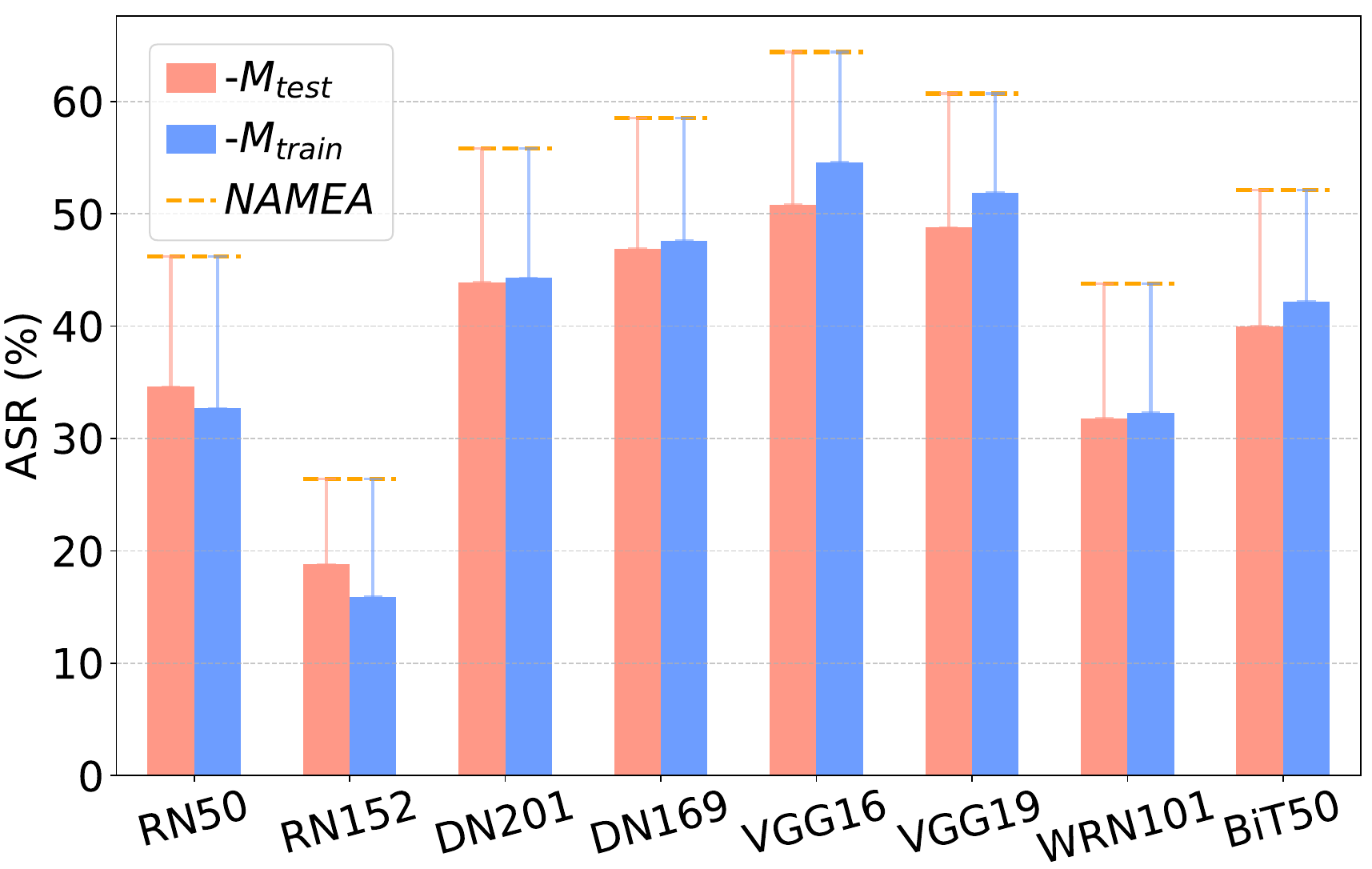}
	\end{minipage}
	\caption{  Ablation study on   gradients respectively from attention  and non-attention areas of ensemble models.
		\textbf{Left:}  ASRs (\%) against ViTs.
		\textbf{Right:} ASRs (\%) against CNNs.
		The base attack is I-FGSM.
	}
	\label{fig:abla_areas}
\end{figure*}

\begin{figure}[!t]
	\centering
	\begin{minipage}[c]{0.23\textwidth}
		\centering
		\includegraphics[width=\linewidth]{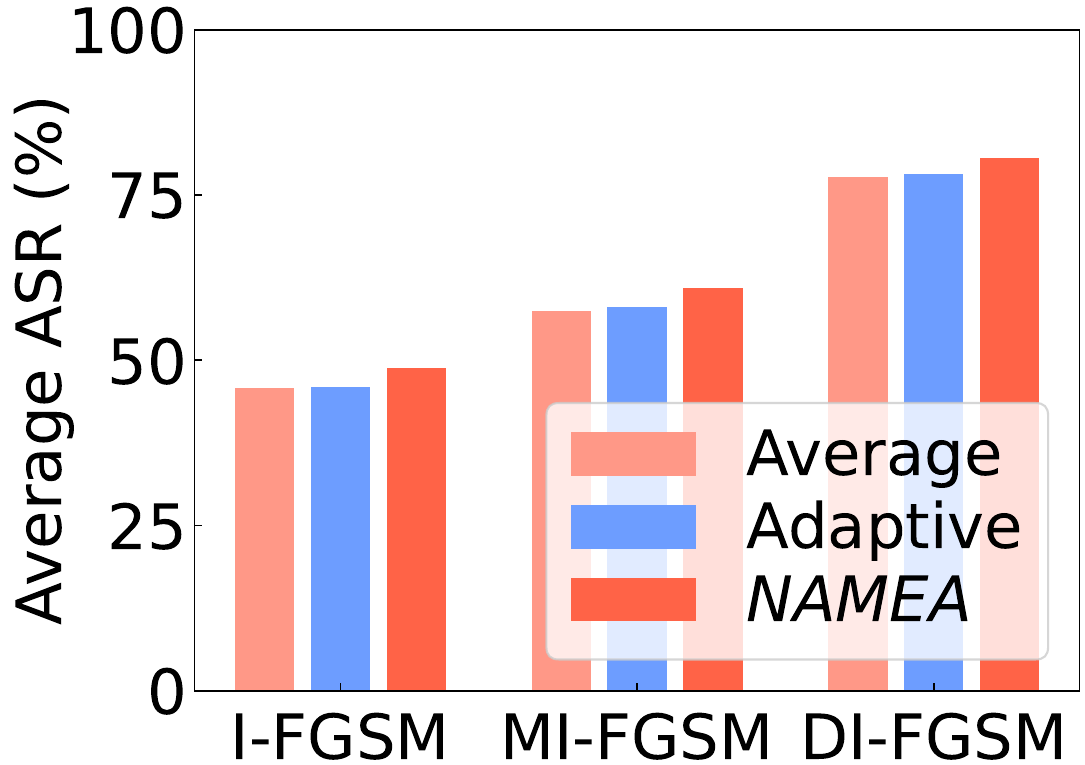}
	\end{minipage}
	\begin{minipage}[c]{0.23\textwidth}
		\centering
		\includegraphics[width=\linewidth]{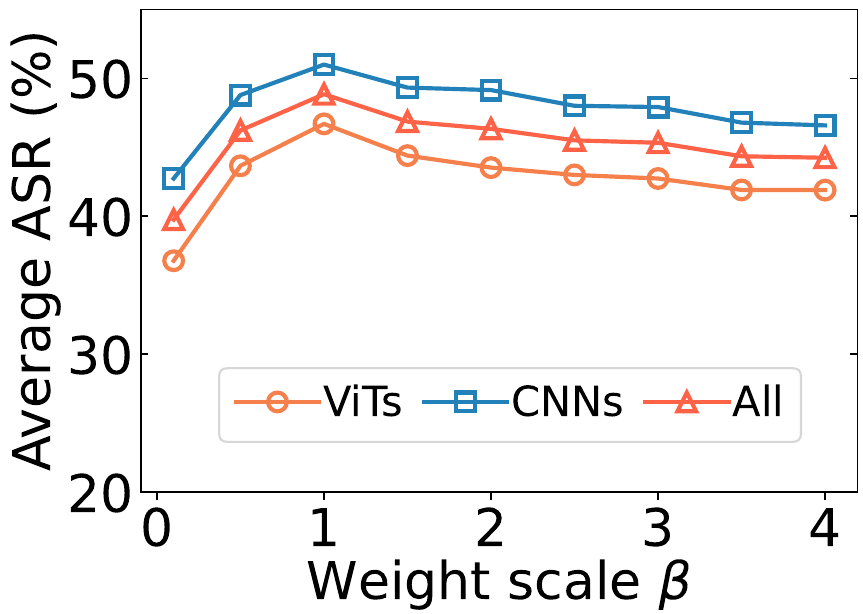}
	\end{minipage}
	\caption{
		\textbf{Left:} Ablation study on gradient aggregation strategies.
		\textbf{Right:} Ablation study on varying weight values.
	}
	\label{fig:abla_area_gradient}
\end{figure}

\begin{figure}[!t]
	\centering
	\begin{minipage}[c]{0.23\textwidth}
		\centering
		\includegraphics[width=\linewidth]{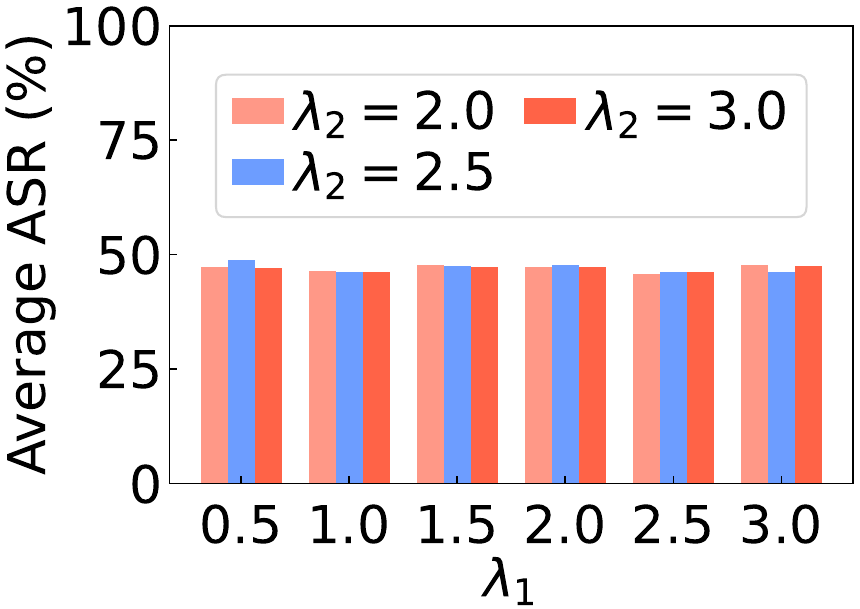}
	\end{minipage}
	\begin{minipage}[c]{0.23\textwidth}
		\centering
		\includegraphics[width=\linewidth]{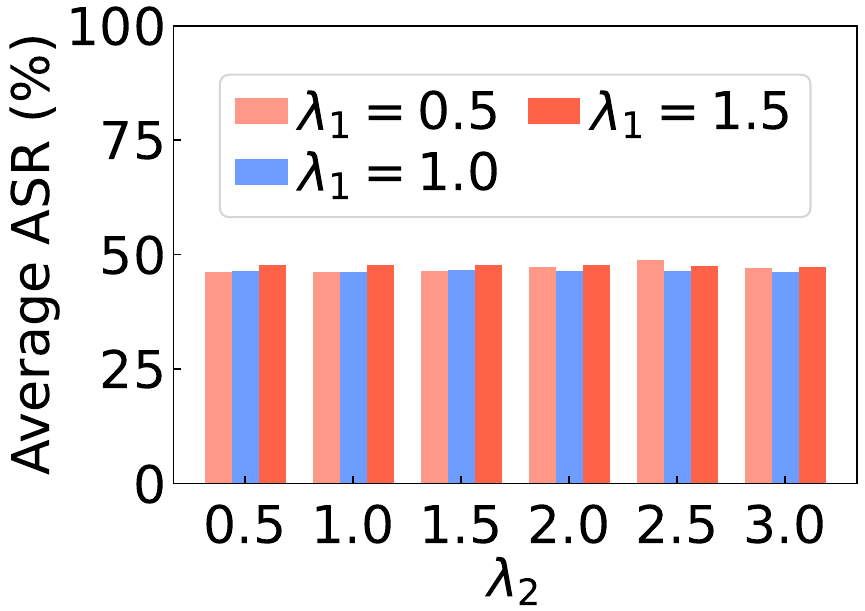}
	\end{minipage}
	\caption{The average ASRs (\%) under  different hyperparameters of  gradient
		scaling for CNNs.}\label{abla_param}
\end{figure}
\textbf{The Effect of Gradients from (Non-)attention Areas.}
\textsf{NAMEA} fuses the gradients calculated  from  attention and non-attention
areas of surrogate models,
which means that the perturbations are added into both areas.
To   evaluate the effect of  the gradients respectively from
attention and non-attention areas,
we design the following ablation settings:
(1)\textbf{-M}$_{test}$ calculates the   final meta gradient as  $g^{t+1} = g^K_{tr}$ to
remove the gradient from non-attention areas;
(2) \textbf{-M}$_{train}$  calculates the   final meta gradient as  $g^{t+1} =  g^K_{te} \odot  \overline{\mathbb{M}}_{K}$ to
remove the gradient from attention areas.
As shown in Fig.~\ref{fig:abla_areas},
both types of gradients exhibit a certain degree of   transferability across CNNs and ViTs,
and fusing them  with meta-learning (i.e., \textsf{NAMEA}) can further enhance   cross-architecture transferability (over 10\% improvement).  


\textbf{The Impact of Gradient Aggregation Strategies.}
According to Eq.~(11) of the main text, the final meta gradient is calculated by  	$g^{t+1} = g^K_{tr} +  g^K_{te} \odot  \overline{\mathbb{M}}_{K}$.
To further validate the effect of our gradient aggregation strategy,
we compare \textsf{NAMEA} with two gradient aggregation strategies: simple averaging (denoted by Average) and reinforcement learning-based adaptive reweighting (denoted by Adaptive).
As shown in the left side of Fig.~\ref{fig:abla_area_gradient}, \textsf{NAMEA} achieves approximately 2.7\% improvement over both strategies.
This improvement may be attributed to the use of non-attention mask in gradient aggregation, which better mitigates gradient interference
between different areas.
Therefore, our meta-gradient aggregation strategy
can  well balance between stable update direction and model diversity.

\textbf{The Impact of Weight of Meta-Testing Gradient.}
According to Eq.~(11) of the main text, the meta-training gradient and the meta-testing gradient are equally important in crafting perturbations.
To   evaluate the effect of two types of gradients in attack performance,
we rewrite Eq.~(11) in a weighted form:
$g^{t+1} = g^K_{tr} +  \beta \cdot g^K_{te} \odot  \overline{\mathbb{M}}_{K}$,
where $\beta$ controls the   relative importance of each type of gradients.
As shown in the right side of Fig.~\ref{fig:abla_area_gradient},
when $\beta = 1$,  \textsf{NANEA}  achieves the best attack performance, and the ASRs against both CNNs and ViTs
show a declining trend as $\beta$  decreases or increases.
This
indicates that meta-training and meta-testing gradients contribute equally to  attack performance.

\textbf{The Impact of Hyperparameters on CNN  Gradient Scaling.}
We investigate the impact of hyperparameters on the performance of layer-wise gradient scaling for CNNs.
According to Eq.~(12) of the main text, both $\lambda_{1}$ and $\lambda_{2}$ jointly influence the degree of gradient scaling.
To evaluate this effect, we conduct ablation experiments under different hyperparameter settings.
As shown in Fig.~\ref{abla_param},   $\lambda_{1}$ and $\lambda_{2}$ has a negligible impact on the performance,
indicating that these hyperparameters are general.
Even with different parameter values, gradient scaling remains effective, making attack performance insensitive to $\lambda_{1}$ and $\lambda_{2}$.

\end{document}